\pdfoutput=1
\documentclass[11pt]{article}

\usepackage{ACL2023}

\usepackage{times}
\usepackage{latexsym}

\usepackage[T1]{fontenc}
\usepackage[utf8]{inputenc}

\usepackage{microtype}

\usepackage{inconsolata}

\usepackage{multirow}
\usepackage{multicol}
\usepackage{amsmath}
\usepackage{amssymb}
\usepackage{graphicx}
\usepackage[linesnumbered,ruled,vlined]{algorithm2e}

\usepackage{booktabs} 
\usepackage{multicol}
\usepackage{multirow}
\usepackage{amsfonts} 
\usepackage{listings}
\usepackage{caption}
\usepackage{xcolor}  % 导入 xcolor 宏包
\usepackage{array}
\usepackage{verbatim}
\usepackage{amsmath}
\usepackage{bbm}
\usepackage{makecell}
\usepackage{pifont}
\usepackage{hyphenat}
\usepackage{xtab,afterpage}

\newcommand{\cmark}{\ding{51}}%
\newcommand{\xmark}{\ding{55}}%
\usepackage{mdframed}
\usepackage{color} 
\usepackage{adjustbox}

\newcommand{\wrong}[1]{{\color{red} #1}}
\newcommand{\avoid}[1]{{\color{brown} #1}}

\usepackage[most]{tcolorbox} 
\newtcolorbox{myquote}[1][]{%
    colback=black!5,
    colframe=black!5,
    notitle,
    sharp corners,
    enhanced,
    breakable,
    }

\usepackage{lipsum}

\title{KnowledGPT: Enhancing Large Language Models with Retrieval and Storage Access on Knowledge Bases }
\author{Xintao Wang, Qianwen Yang, Yongting Qiu, Jiaqing Liang, \\ \textbf{Qianyu He,  Zhouhong Gu, Yanghua Xiao, Wei Wang} \\
Fudan University \\ 
\texttt{\{xtwang21, qwyang22, 22210240256, qyhe21\}@m.fudan.edu.cn,} \\ \texttt{\{zhgu20, shawyh, weiwang1\}@m.fudan.edu.cn,} \texttt{l.j.q.light@gmail.com}
}

\begin{document}
\maketitle
\begin{abstract}
Large language models (LLMs) have demonstrated impressive impact in the field of natural language processing, but they still struggle with  several  issues regarding, such as completeness, timeliness, faithfulness and adaptability. 
While recent efforts have focuses on connecting LLMs with external knowledge sources,  the integration of knowledge bases (KBs) remains understudied and faces several challenges. 
In this paper, we introduce KnowledGPT, a comprehensive framework to bridge LLMs with various knowledge bases, facilitating both the  retrieval and storage of knowledge.
The retrieval process employs the program of thought prompting, which generates search language for KBs in code format with pre-defined functions for KB operations. 
Besides retrieval, KnowledGPT offers the capability to store knowledge in a personalized KB, catering to individual user  demands. 
With extensive experiments, we show that by integrating LLMs with KBs, KnowledGPT properly answers a broader range of questions requiring world knowledge compared with vanilla LLMs, utilizing both knowledge existing in widely-known KBs and extracted into personalized KBs. 
%With both qualitative and quantitative experiments, we show that integrating LLMs and KBs is a promising way. 
\end{abstract}
%from widely-known KBs and knowledge storage into personalized KBs.
%is also equiped with a personalized KB, that supports both read and write operation on various types of knowledge.  

\section{Introduction}

% 语言模型 prevailing
% Large Language Models  (LLMs) have been prevailing in NLP tasks...
Large Language Models (LLMs)~\citep{openai2023gpt4} ~\citep{vicuna2023}~\citep{taori2023alpaca} have achieved substantial impact across a variety of natural language processing (NLP) tasks like translation~\citep{wang2023document}, summarization~\citep{zhang2023benchmarking} and question answering~\citep{van2023open}, alongside with all kinds of requests from real-world users. 
% 各种各样的user requests
%Furthermore, they are also proficient in answering all kinds of real-world user requests.
% 源于数据量大
Their remarkable capabilities stem from the ever-increasing number of  parameters and training data, which expands in correlation with their massive knowledge and emergent abilities like chain-of-thought reasoning~\citep{kojima2022cot} and in-context learning~\citep{gpt3}.

\begin{figure}[t]
    \centering
        \includegraphics[width=1\linewidth]{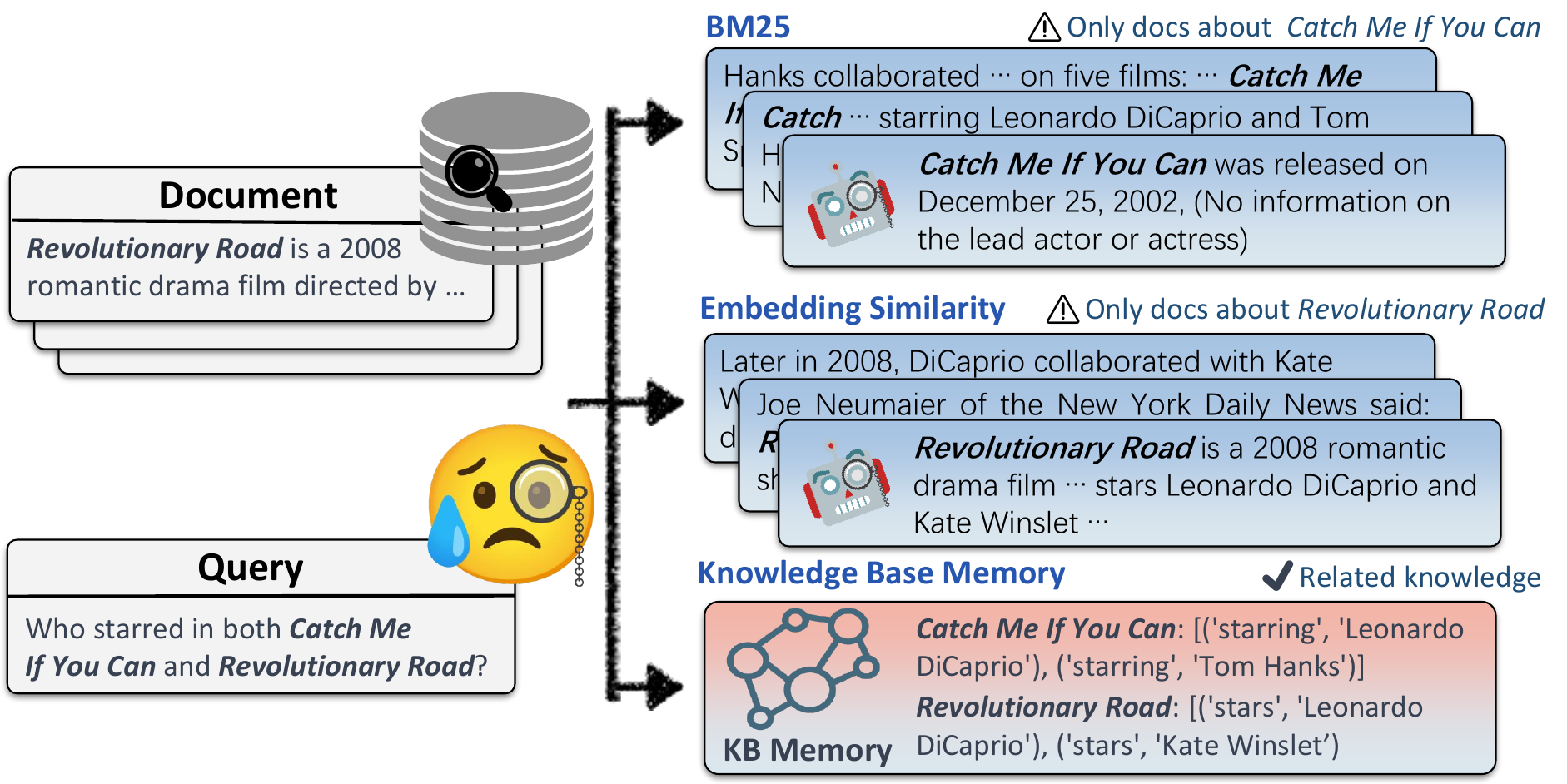} 
    \caption{Comparison between retrieval results from document corpus and knowledge bases. In this case,  no document retrieved from the corpus provides enough knowledge for the query, while
    from the knowledge base sufficient related knowledge can be retrieved.  }
    \label{fig:why_kggpt}
\end{figure} 

% Despite their ..., LLMs still suffers from several problems about knowledge. 
However, LLMs still struggle with factual knowledge considering issues like completeness, timeliness, faithfulness and adaptability~\citep{openai2023gpt4}~\citep{kandpal2023large}.
%First, LLMs still lack of certain knowledge, especially latest world knowledge,  domain-specific or personalized knowledge. 
Firstly, LLMs demonstrate limitations in terms of timely updates and expertise in specific domains.     
%(2) hallucination, generate unfaithful knowledge ...  
Secondly, these models can generate unfaithful or  ``hallucinated''  knowledge, posing both reliability and ethical concerns.
% (3) Considering that LLMs can hardly be tuned to incorporate new knowledge 
Thirdly, due to constraints like costs and accessibility, LLMs can hardly incorporate new knowledge via continued training, which hampers the ability to tailor these models to accommodate specific knowledge demands. 
% summarize
Therefore, these knowledge demands encourage comprehensive research efforts towards integrating LLMs with external sources of knowledge.

% 解决知识问题 - 接入知识库
% done ljq0628: 2种逻辑：1 同时写接KG的和接向量数据库的； 2 第二段再写 除此之外 还有向量数据库的
%%% Addressing these challenges has recently seen a shift towards equipping LLMs with the capacity to utilize plugins, tools, and memories to enhance knowledge retrieval.
Towards this issue, some recent efforts have been made to enable LLMs to access plug-and-play knowledge sources like knowledge bases (KBs)~\citep{modarressi2023retllm}, search engines~\citep{schick2023toolformer}, document memories~\citep{Liu_LlamaIndex_2022}, and databases~\citep{hu2023chatdb} to provide LLMs with world knowledge, generally via LLM-generated API calls.  
% 我们关注KG
In this paper, we focus on knowledge bases (KBs), a unique form of knowledge source featuring entity-centric knowledge like relational triples and entity descriptions. 
% KG important: explainablity
On one hand, various KBs have been constructed for their practical effectiveness for applications, and the conciseness, expressiveness, interpretability and visibility of their representation. 
% KB for LLM 有其独特的应用场景，与文档检索不同，现有文档方法不够用
On the other hand, previous approaches have largely focused on document corpus, which reveals several deficiency when applied to KGs, as shown in Fig \ref{fig:why_kggpt}.
% Beyond, KG for LLM 依然有其重要性，.
Therefore, connecting LLMs with KBs is of significant importance, yet still remains underexplored.

% 现有工作
Recently, several works have attempted to connect LLMs to KBs.
% Toolformer
Toolformer~\citep{schick2023toolformer} queries Wikipedia for descriptions of interested entities to answer related  questions. 
% Graphtoolformer
Graph-Toolformer~\citep{zhang2023graphtoolformer} and ToolkenGPT~\citep{hao2023toolkengpt} make LLMs reason over knowledge graphs like Freebase.
% personal
RET-LLM~\citep{modarressi2023retllm} builds personalized KG memory with relational triples extracted from past conversations for future use, in parallel with practical  efforts of KG Index in LangChain~\citep{Chase_LangChain_2022} and Llama Index~\citep{Liu_LlamaIndex_2022}.

\begin{figure*}[htbp]
    \centering
        \includegraphics[width=1\linewidth]{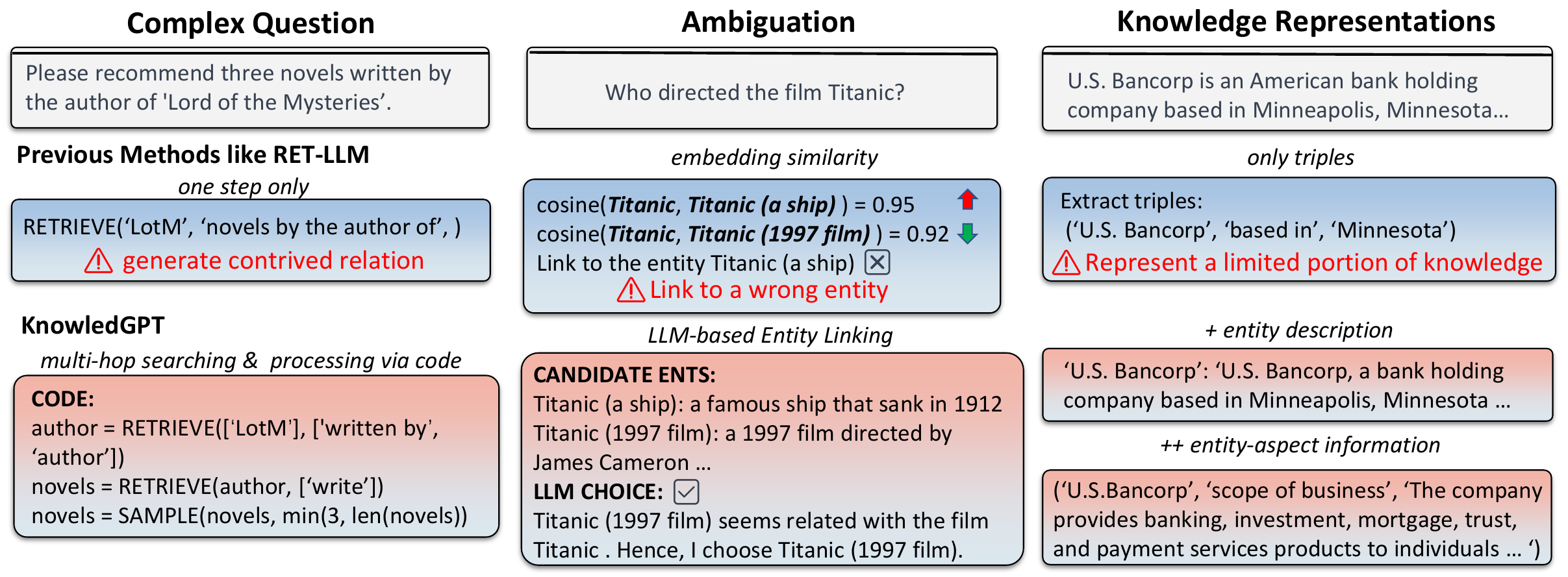} 
    \caption{Comparison between KnowledGPT and previous methods towards several challenges in bridging LLMs with KBs. KnowledGPT handles complex questions through multi-hop searching and processing based on code, tackles entity ambiguation with LLM-based entity linking (middle), and provides extended forms of knowledge representations to encapsulate a broader range of knowledge from the provided text. }
    \label{fig:challenges}
\end{figure*}  

However, there are still many challenges in this direction, as shown in Figure~\ref{fig:challenges}. 
% Frist, How LLMs search over knowledge bases for diversified and complex queries? There may be multi-hop questions. 
% ljq: 加个例子；这一段是最难的，别人不能做的，我们能做的；然后用例子做论据。
Firstly, the process by which LLMs navigate through knowledge bases for intricate and varied questions remains a problem, especially for multi-hop questions which requires  information across multiple and nested entries in KBs.
% Second, how to align entities and relations in knowledge bases, considering both their diversified expressions in natural language and frequent ambiguation in knowledge bases. 
Secondly, aligning entities and relations in knowledge bases with their text mentions is a challenging task, as they need to map to a wide spectrum of natural language expressions and account for severe ambiguation in the knowledge bases. 
% Third, what knowledge representations should be considered by symbolic memory that are most suitable for LLMs to extract and retrieve, beyond merely triples? Triples can only express limited knowledge compared with language ... 
% ljq: 不能是建议 得是个强硬的东西，三元组能表示的知识有限；
Thirdly, while the triple-based representation in KGs is neat and interpretable, it only covers limited information compared with natural language, which suggests the need for new representation form in KBs for LLMs. 
In this paper, we propose a comprehensive framework, KnowledGPT, to connect LLMs to various knowledge bases effectively, with an improved capability in dealing with complex questions, ambiguation and knowledge representation.
KnowledGPT implements a unified accessor interface for operations on different KBs, including widely-used public KBs and personalized KB memories. 
% KnowledGPT accesses entity-oriented knowledge, including entity description, relational triples, as well as a combination of that which we name entity aspect information.  
KnowledGPT accesses entity-oriented knowledge, including both entity descriptions and relational triples.
% As for searching, to answer a given query, KnowledGPT uses three steps: search code generation, search execution and answer generation. 
For a given query, KnowledGPT searches KBs with three steps: search code generation, search execution, and answer generation.
% The search code generation step prompts GPT-4 to generate search language for KBs in the form of codes by Program of Thought (PoT) prompting. 
% The search execution step executes the search code with a unified KB accessor on multiple KBs in parallel. This includes an entity linking step based on GPT-4 to avoid ambiguation. 
% The answer generation step finally integrates the retrieved knowledge and the query to give an informed answer. 
% Besides searching, KnowledGPT can also extract knowledge from unstructured texts to populate the personalized KB. 
% Different from some recent works like GPT-Index, LangChain KB Memory and RET-LLM that extracts only relational triples, we extract entity-oriented knowledge in three forms of representations, including entity description, relational triples, as well as a combination of that which we name entity aspect information.
% This enables our personalized KB to represent more kinds of world knowledge.  
Inspired by ~\citep{chen2022pot}, KnowledGPT adopts \textit{program of thoughts (PoT)} prompting, interacting with KBs by generating Python code which delegates searching steps and executing it. 
This code encapsulates functions for assessing KBs such as \textit{entity\_linking}. 
Afterwards, KnowledGPT integrates the retrieved knowledge to generate the response. 
If KnowledGPT judges that the question does not necessitate knowledge from KBs, or if the retrieved knowledge is inadequate or absent, the question will be directly answered by the LLM. 
%The search execution step executes the search code using the unified KB accessor on multiple KBs in parallel, which includes an entity linking step based on GPT-4 to mitigate issues arising from ambiguation.
%In the final answer generation step, the retrieved knowledge is integrated with the query to produce an informed response.
Besides, KnowledGPT can also extract knowledge from unstructured texts represented in various forms to enrich the personalized KB. 
%Unlike some recent works such as GPT-Index, LangChain KB Memory, and RET-LLM which focus on extracting relational triples, we extract entity-oriented knowledge in three forms of representations, furthering the ability of our personalized KB to represent a wider variety of world knowledge.

%To summarized, our contributions are as follows:
% First, we propose a comprehensive framework to enable LLMs to retrieve knowledge from KBs, which provides solutions to pratical application issues. 
% Second, we propose to use personalized KBs as symbolic memory, with three types of entity-oriented knowledge representations. 
% Third, we show the effectiveness of our methods with detailed experiments. 

Overall, our contributions are summarized as follows:
\begin{enumerate}
    \item We propose KnowledGPT, a comprehensive framework to enable LLMs to retrieve knowledge from knowledge bases. 
    It significantly advances the collaboration between  LLMs and KBs towards vital practical challenges like complex  searching and ambiguation.
    \item We propose the use of personalized knowledge bases as  symbolic memory for LLMs, encapsulating entity-oriented knowledge in three forms of representations. 
    This expands the scope of knowledge in symbolic memories compared with triples-only KBs. 
    %accessed and used by LLMs.
    %with various forms of representations. This symbolic memory encapsulates entity-oriented knowledge in three forms of representation, significantly expanding the scope and type of knowledge that can be effectively accessed and used by LLMs.
    \item We demonstrate the efficacy of our proposed methods with  experiments. The results underscore the utility and potential of using KBs as symbolic memory for LLMs. 
    %our framework in improving the performance of LLMs across a range of tasks and scenarios.    
\end{enumerate}

% 0809: 产生副结论：在QA里，多大的模型能产生什么效果。
% 0809: 加个vicuna试试 - 还得压缩prompt，prompt长超过2k; done 
% challenge: 实验数据太小 / 缺乏对比模型 / 如何和document QA 比较 / 小模型不适用 / 效率有问题 ；找个借口将审稿人堵上
\section{Related Works}
\paragraph{\textbf{External Knowledge and Memory for LLMs}}
% LLMs strong
Large language models (LLMs), such as GPT-4~\citep{openai2023gpt4} and LLaMA~\citep{touvron2023llama}, have demonstrated impressive performance across various applications. 
% knowledge
However, they still struggle with knowledge considering completeness, timeliness, faithfulness and adaptability. 
% augment
Hence, many recent efforts have been devoted to equipping LLMs with external knowledge.
% search engine
Internet-augmented language models ~\citep{komeili-etal-2022-internet}~\citep{lazaridou2022internet},
as well as New Bing and \texttt{ChatGPT} ``Browse with Bing''  plugin, allow LLMs to access up-to-date information with search engines or web browsers.
% doc base 
Retrieval-augmented methods like REALM~\citep{guu2020retrieval}, RAG~\citep{lewis2020retrieval}, augment LLMs with document corpus, which have also been increasingly adopted by recent popular LLMs like \texttt{ChatGPT}
as the memory unit~\citep{Liu_LlamaIndex_2022}~\citep{Chase_LangChain_2022}.
%ChatDB
ChatDB~\citep{hu2023chatdb} augments LLMs with databases as symbolic memory. 

\paragraph{\textbf{Knowledge Bases for LLMs}}
Some recent works have studied to augment LLMs with knowledge from external KBs or use KBs as symbolic memories, usually by making LLMs generate API calls for KB operations. 
% Tool-former
Toolformer~\citep{schick2023toolformer} trains LLMs to search Wikipedia for texts of entities. 
% Graph-toolformer
Graph-Toolformer~\citep{zhang2023graphtoolformer} empowers LLMs to reason over knowledge graphs. However, it skips the entity linking step, so it requires  entity id like /m/053yx as input, instead of their names.
% toolkenGPT
ToolkenGPT~\citep{hao2023toolkengpt} keeps the LLMs frozen and  trains tool embeddings for relations in KBs to support relational queries.
% RET-LLM ... 
RET-LLM~\citep{modarressi2023retllm}, similar to the KG memory of LangChain~\citep{Chase_LangChain_2022} and Llama-Index~\citep{Liu_LlamaIndex_2022}, extracts relational triples from user inputs and store them in a symbolic KG memory. 
Compared with previous efforts, KnowledGPT supports various knowledge representations and both public and private KBs,  as shown in Table ~\ref{tab:comparison}.
% our method - difference
%The difference between existing methods and our KnowledGPT are shown in Table ~\ref{tab:comparison}.

% RET-LLM: 实现了私有库的存取。私有库仅限三元组。
% similar to toolformer, RET-LLM tunes a LLM with synthetic dataset

\begin{table*}
\setlength\tabcolsep{3pt}
\centering
\resizebox{\textwidth}{!}{
\begin{tabular}{lcccccccc}
    \Xhline{0.8pt}
    %\multirow{2}{*}{\multicolumn{1}{l}{\textbf{Method}}}  & \multicolumn{3}{c}{Type of Knowledge} & \multicolumn{3}{c}{ConcEPT} \\
    %\cmidrule(lr){3-5} \cmidrule(lr){6-8}
    %& & Score & & Similarity & Score &&  Similarity \\
    Method & Ent Desc & Triples & Ent Aspect Info & Multi-Hop Search & Zero-Shot & Ent Linking & External KBs & Private KB 
    \\
    \hline
    %{\xmark} & {\cmark} \\ \hline
    Toolformer & \cmark & \xmark & \xmark & \cmark & \xmark & \xmark & \cmark & \xmark  \\
    ToolkenGPT & \xmark & \cmark & \xmark & \cmark & \xmark & \xmark & \cmark & \xmark \\
    Graph-Toolformer & \xmark & \cmark & \xmark & \xmark & \xmark  & \xmark & \cmark & \xmark \\
    RET-LLM & \xmark & \cmark & \xmark & \xmark & \xmark  & \cmark & \xmark & \cmark \\
    LangChain KG Memory & \xmark & \cmark & \xmark & \xmark & \cmark  & \cmark & \xmark & \cmark \\
    Llama-Index KG Index & \xmark & \cmark & \xmark & \xmark & \cmark  & \cmark & \xmark & \cmark \\
    %KBQA Models & \xmark & \cmark & \xmark & \cmark & \xmark & \cmark & \cmark & \cmark \\
    KnowledGPT & \cmark & \cmark & \cmark & \cmark & \cmark & \cmark & \cmark & \cmark \\
    \Xhline{0.8pt}
    
\end{tabular}}
\caption{
Comparison between  KnowledGPT and existing KB-augmented methods. Ent, Rel, and Desc are abbreviations for entity, relation and description, respectively. 
}
\label{tab:comparison}
\end{table*} 

\paragraph{\textbf{Knowledge-based Question Answering}} 
(KBQA) is to search for answer entities or relations  
given natural language queries specific to certain KGs. 
Existing KBQA systems are mainly based on semantic parsing~\citep{Berant2013SemanticPO} or information extraction~\citep{yao2014information}, where language models are increasingly involved.
% semantic parsing
Semantic parsing methods~\citep{yu2023decaf, cao2022program, zhang2022crake, Abdelaziz_Ravishankar_Kapanipathi_Roukos_Gray_2021, lai2016open} leverage semantic parser to convert natural language queries into intermediate logic forms such as SPARQL~\citep{sparql} and program~\citep{liang2016neural}, which are executed on KBs to obtain the answers. 
However, the generated logic forms are usually non-executable, thus  failing to arrive at the correct answer~\citep{sun2020sparqa}. 
Pangu~\citep{gu2022pangu} trains a language model discriminator to evaluate probability of candidate plans. 
% information extraction
Information extraction methods usually combines retrieval and reasoning~\citep{Zhang_2022, shi2021transfernet, sun2019pullnet, jiang2023unikgqa, baek2023knowledge}. 
%It is easy to handle single-hop question, but the number of multi-hop relations grow explosively with the increase of hops, which is intractable for both storage and computation~\citep{zhang2017variational}.
These methods show effectiveness in handling single-hop retrieval. 
However, they encounter challenges with multi-hop retrieval concerning storage and computation costs, where the number of relations expands exponentially with each added hop. %This increase becomes computationally intractable, presenting substantial issues for both storage and computation.

% RNG-KBQA~\citep{ye2022rngkbqa} uses a generation model conditioned on the question and the top-ranked candidate logical forms to compose the final logical form executed against the KBs.
%RNG-KBQA~\citep{ye2022rngkbqa} uses a generation model and top-ranked candidates to compose the final logical form executed against the KBs.
% Program induction for answering complex questions over KBs aims to decompose a question into a multi-step program. 
%Program transfer~\citep{cao2022program} leverage program annotations on the rich-resourced KBs as external signals to aid program induction for the low-resourced KBs.
%UniKGQA~\citep{jiang2023unikgqa} utilizes a semantic matching module and a propagation module to propagate the matching information along the directed edges on KBs.
KnowledGPT differs from KBQA methods in two aspects.
First, many KBQA methods are designed for special queries about relational triples in KGs, 
%aim to locate entities or relations for special queries, 
while KnowledGPT augments LLMs to respond to various user queries with knowledge in various forms from KBs.  
Second, KBQA methods are typically trained on specific datasets and KGs, whereas KnowledGPT requires no training and can easily accommodate different LLMs and KBs.

% 这段表明现有KBQA方法只能适应格式较为固定的、特定于三元组的问句，而没法处理真实世界中的各种问题。
%traditional KBQA only handles simple questions like 'who is the founder of Apple Inc.', and struggles with complex and diverse user inputs like 'Who will win if Socrates fight with Plato', 'write a resume for Plato' and 'Which anime has the same voice actors for the male and female leads as "Sword Art Online"?'

%These KBQA methods generally needs training on specific datasets. 
%KnowledGPT don't train. 

\section{Methods} 
In this section, we introduce KnowledGPT, a comprehensive framework to integrate LLMs with KBs. 
We first provide the definition of two tasks of KnowledGPT,  knowledge retrieval and knowledge storage (Sec ~\ref{sec:taskdefinition}). 
Then, we elaborate the details in the retrieval (Sec ~\ref{sec:retrieval}) and storage (Sec~\ref{sec:storage}) process of KnowledGPT. 

%Then, we present the framework of how KnowledGPT retrieves knowledge from KBs~\ref{sec:retrieval}. 
%Afterwards, we introduce the details of KnowledGPT~\ref{sec:storage}.

\subsection{Task Definition}
\label{sec:taskdefinition}
% kbs
KnowledGPT supplements LLMs with external knowledge from various knowledge bases (KBs), including a personalized KB (PKB) as an writable symbolic memory.
% KnowledGPT有两个任务: 查询和增加
Given a user input in natural language, KnowledGPT undertakes two primary tasks, namely knowledge retrieval and knowledge storage. 
% Search
In the knowledge retrieval task, the model searches through provided KBs to retrieve relevant knowledge to answer the user query.
% Add 
In the knowledge storage  task, the model extracts knowledge from the user input and inserts it into the PKB. 

\subsection{Knowledge Retrieval}
\label{sec:retrieval}
\begin{figure}[tb]
    \centering
        \includegraphics[width=1\linewidth]{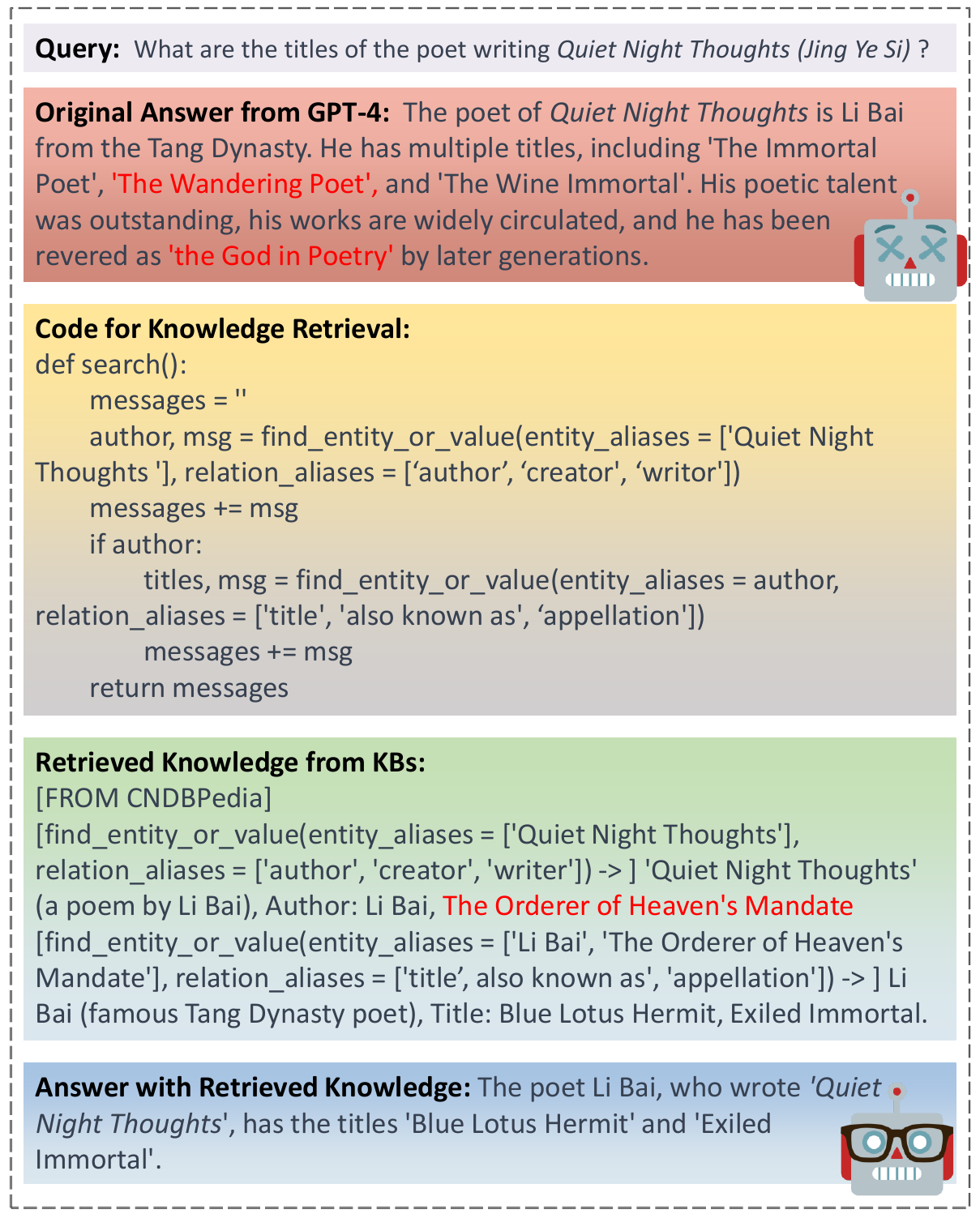} 
    \caption{
    The detailed process of KnowledGPT's  retrieval steps 
    for the given example. 
    It shows the generated code for retrieval, the retrieval results,  and the knowledge-enhanced response for the query, along with the original answer from LLMs. Texts highlighted in red indicates inaccuracies. }
    \label{fig:full_example}
\end{figure}

% overall framework
KnowledGPT follows a three-step process to answer user queries with knowledge from KBs, as shown in Fig ~\ref{fig:full_example}. 
First, it generates a piece of search code as a logical form for query-specific KB access. 
Then, the search code is executed to retrieve related knowledge. 
Finally, KnowledGPT reads the retrieved knowledge and answers the query.

% 人类请求是多样复杂的
%In practical applications of LLMs, the user inputs cover a wide range of complexity and diversity, such as \wxt{'List the protagonists of the four great tragedies.'}.
%This makes it difficult to retrieve the needed knowledge, which exceeds the capacity of existing KBQA methods~\citep{}~\citep{}.

%\subsubsection{Search Code Generation}

% 我们用LLM的代码生成解决问题 PoT
We utilize the \textit{program of thought (PoT)} prompting approach~\citep{chen2022pot} , which adopts Python code as the search language generated by LLMs. 
In this paper, we use \texttt{GPT-4}~\citep{openai2023gpt4} as the LLM. 
The code is encapsulated in a \textit{search} function, 
as is shown in the yellow part of Fig ~\ref{fig:full_example}
, which includes 
% unified interface, three functions
built-in Python functions and  three custom KB functions designed to facilitate the interaction of LLMs with KBs:
\begin{enumerate}
    \item \textit{get\_entity\_info}, which accepts an entity as input and returns its encyclopedic description.  
    \item \textit{find\_entity\_or\_value}, which accepts a query consisting of an entity and a relation as input, and outputs a list of the corresponding entity or value. 
    \item \textit{find\_relationship}, which accepts two entities as input, and returns a list of their relationship. 
\end{enumerate}
Specially, each entity or relation is represented as a list of candidate aliases, rather than a single name, to effectively handle synonyms. 
Besides the outputs stated above, these KB functions also return a message which logs the function call and result. 
Then, the overall output of the \textit{search} function is obtained by concatenating the messages from individual KB function calls. 
The prompt is shown in ~\ref{fig:prompt_search}. 

%\wxt{examples of functions input and output are shown in figure}
%For instance, given the input 'Socrates', the function outputs 'Socrates is...'.
%E.g. Input: ('The Republic', 'author'), Output: 'Plato'.
%E.g. Input: ('Bill Gates', 'Microsoft'), Output: 'founder of'

The \textit{search} function is then executed to retrieve the expected knowledge from KBs.
% decorate code
The code would be decorated before execution, e.g. with a try-except statement and KB-specific accessor object, as is elaborated in Sec ~\ref{sec:execution}.
% multiple kbs 
The \textit{search} function is executed for each KB respectively in parallel, and their results are concatenated.  

Finally, the retrieved knowledge is provided to LLMs, and LLMs are tasked with responding to the user's query,  supported by  the retrieved knowledge. The prompt is shown in Sec ~\ref{fig:prompt_search}.
LLMs will ignore the retrieved information and address the user query independently in scenarios where LLMs judge the question does not require external knowledge or the retrieved knowledge is not enough for the query.

%\begin{figure*}[htbp]
%    \centering
%        \includegraphics[width=1\linewidth]{pics/whiteboard_exported_image.png} 
%    \caption{Our framework.}
%    \label{fig:framework}
%\end{figure*}  

\subsubsection{Code Implementation}
\label{sec:execution}

Next, we introduce the implementation of the KB functions to execute the generated code. 
% what we implement %design a unified accessor interface for different KBs. 
We implement the functions  at two levels: a unified level and a KB-specific level.

Functions at the unified level provide a unified interface for operations over different KBs. These include the three KB functions (\textit{get\_entity\_info}, \textit{find\_entity\_or\_value}, \textit{find\_relationship})  generated directly by LLMs, and an \textit{entity\_linking} function to align entity aliases generated by LLMs with entities in KBs. 

Functions at the KB-specific level implement operations on each specific KB by calling corresponding APIs. Basically, we only need to implement three functions for each KB: \textit{\_get\_entity\_info}, \textit{\_entity\_linking}, \textit{\_get\_entity\_triples}. We denote these functions with an underscore in front in this paper. 

% two level, unified level / KB-specific level
%Basically, we implement the functions at two levels: a unified level which unifies operations over different KBs, and an inner kb-specific level which calls api of each kb (starting with \_ in this paper).

% try except
Prior to execution, we decorate the generated code. 
We wrap the code with a try-except statement, so that if the code breaks down in subsequent steps, the \textit{search} function  still returns valuable results from successful steps. 
Also, we  pass the user query into the \textit{search} function as a global variable.

%\wxt{an example of actual code implemented}

\subsubsection{Entity Linking}

\begin{figure}[htbp]
    \centering
        \includegraphics[width=1\linewidth]{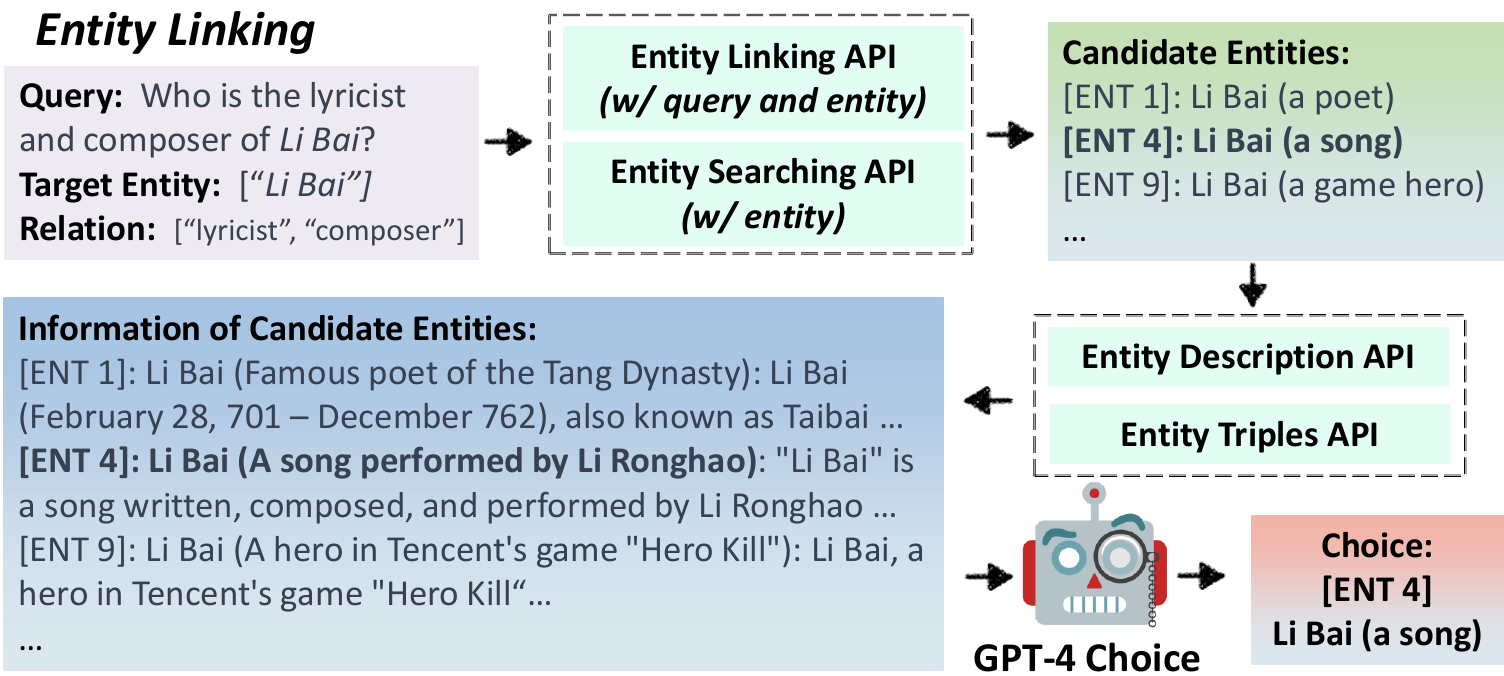} 
    \caption{The detailed process of KnowledGPT's entity linking steps for the given example. }
    \label{fig:el_process}
\end{figure}

% aliases
%Note that, for all these functions, the entity/relation input is not a single string, but a list of its aliases generated by GPT-4. 
%The \textit{entity\_linking} functions would traverse entity aliases, and link to an appropriate one. 

% What is el
Entity linking, which aligns entity mentions in natural language with entities in a KB, is an indispensable step towards the integration of LLMs with KBs. 
It is essential because an entity can be referred to by various mentions (e.g. \textit{Donald Trump} and \textit{President Trump}), and a noun phrase can also refer to different entities (e.g. the fruit \textit{apple} and the company \textit{Apple}).

% how: for public KBs
%For public KBs, 
Our \textit{entity\_linking} function comprises three steps, as depicted in Fig ~\ref{fig:el_process}. 
First, we invoke the KB-specific \textit{\_entity\_linking} function  to obtain candidate entities. 
% inner level
It basically takes as input the query and entity aliases, and utilizes the entity linking API (with both entity names and the context) and searching API (with only entity names) provided by the corresponding KB.  
% private kb
%\wxt{In cases where no such API is available, such as for PKB, we can resort to simple string matching and cosine similarity using openai text-embedding-ada-002 model as an alternative approach.} %, and each entity have multiple mentions. \footnote{} 
Second, we call the \textit{\_get\_entity\_info} function (introduced in Sec. ~\ref{sec:get_entity_info}) to gather information about the candidates. 
Each piece of entity information will be truncated to a maximum length. 
Finally, we provide LLMs with the function input (including the query, aliases of entity and relation) and  the candidate entities along with their information, and make LLMs determine the most appropriate entity.

% for private KG
%For relations, we align relation aliases generated by LLMs with relations in the KB by similarity measures. Specifically, we compute cosine similarity between the sentence embeddings representing the relations. 

%\begin{figure*}[htbp]
%    \centering
%        \includegraphics[width=1\linewidth]{pics/whiteboard_exported_image (1).png} 
%    \caption{Our framework.}
%    \label{fig:framework}
%\end{figure*}  

\subsubsection{Get Entity Information}
\label{sec:get_entity_info}

The \textit{get\_entity\_info} function retrieves information about a specific entity. 
It first employs the \textit{entity\_linking} function to link entity aliases to an entity in the KB. 
Subsequently, it invokes the KB-specific \textit{\_get\_entity\_info} function, which returns information of the given entity in KB, including its entity description and triple information. 
The \textit{\_get\_entity\_triples} function is called to collect its triples.
% inner function called by find_entity_or_value or find_relationship
The KB-specific \textit{\_get\_entity\_info} function is nested in the \textit{entity\_linking} function, which makes it an integral part of all KB functions at the unified level. 
%When nested in the \textit{find\_entity\_or\_value} function, it additionally considers relation aliases as input and prioritizes triples of similar relations for improved entity linking results. 

%priority to similar relations based on Jaccard distance, ensuring more accurate results during the retrieval process.
%also take as input the relation aliases, and it will prioritize similar relations (judged by jaccard distance).

\subsubsection{Find Entity or Value}

% input, output, define task
Given a query composed of an entity and a relation, the \textit{find\_entity\_or\_value} function is designed to retrieve the corresponding entity or attribute value. 
This function undergoes several steps, as is presented in Algorithm  ~\ref{alg:find_entity_or_value}.
% steps:
Still, it starts by invoking the \textit{entity\_linking} function to associate entity aliases with a corresponding entity in the KB. 
Then, it calls the internal \textit{\_find\_entity\_or\_value} function, which includes a KB-specific \textit{\_get\_entity\_triples} functions that retrieve all triples related to the entity.
Subsequently, the relations in these triples are sorted based on their similarity with the input relation aliases. 
Here we employ cosine similarity of sentence embeddings, rather than symbolic metrics, which considers synonyms of relations. 
% and embeddings of GPT-3 Ada
%to encode relation strings considering a relation can be represented by synonyms, and calculate their cosine similarity.}
Afterwards, we select the relation with highest similarity score, and return entities or attribute values from all corresponding triples. 
%\wxt{
To improve the robustness of our method, we will conduct a further search within the entity description for the relation if no triples are found. 
If the relation is present in the description, we return the corresponding sentence.
Otherwise, we return the whole description, which may still offer relevant details for LLMs.
%}

\begin{algorithm}[t]
%   \normalem
   \caption{find\_entity\_or\_value}
   \label{alg:find_entity_or_value}
%   \begin{small}
   \begin{footnotesize}
%   \BlankLine
    %\KwIn{$query, entity\_aliases, relation\_aliases$}
   \KwIn{query $q$, alias list of entity $\mathcal E$, alias list of  relation $\mathcal{R}$.}
   \KwOut{a list of target entities or attribute values $\mathcal{T}$.}
    \DontPrintSemicolon
    \SetKwFunction{FMain}{EMBSIM}
    \SetKwProg{Fn}{Function}{:}{}
    \Fn{\FMain{str $r$, list[str] $\mathcal{R}$}}{
        $s=-1$\;
        $\textbf{r}=\texttt{embedding}(r)$\;
        \For{$r_i \in \mathcal{R}$}{
            $\textbf{r}_i=\texttt{embedding}(r_i)$\;
        \If{\texttt{cos}$(\textbf{r}, \textbf{r}_i) > s$}{
            $s=\texttt{cos}(\textbf{r}, \textbf{r}_i)$
        }}
        
        \KwRet $s$\;
    }
    $e=$\texttt{entity\_linking}$(\mathcal E)$\;
    \uIf {$e == NULL$} {
            \Return $NULL$\;
    }
    $r = NULL$\;
    $s_{r} = -1$\;
    \For{ $triple \in triples$}{
        $r_i=triple.rel$\;
        $s_i=\texttt{EMBSIM}(r_i, \mathcal{R})$\;
        \If{$ s_i > s_r$ }{
            $s_r$ = $s_i$\;
            $r = r_i$
        }
    }
    \uIf {$ r == NULL$} {
            \Return $NULL$\;
    }
    $triples_{r}$ = $triples$ with relation $r$\;
    $\mathcal{R}=$ target entities or attribute values in $triples_{r}$\;  

\KwRet \texttt{$\mathcal{R}$}\;
% \vspace{-2mm}
\end{footnotesize}
\end{algorithm}

\subsubsection{Find Relationship}
%\wxt{use psuedo code to replace figure 4}

Given a query composed of two entities, the \textit{find\_relationship} function is designed to retrieve their relationship. 
This functions is similar to \textit{find\_entity\_or\_value}. 
The difference is that, upon retrieving triples 
or entity information
for the first entity, the \textit{find\_relationship} function proceeds to search for  the second entity, instead of the relation. 
If this initial search fails, the function swaps the first entity and second entity and searches again. 
Different from relation similarity, we measure entity similarity by Levenshitein distance $d$. The entity similarity is calculated as $100-d$ if two entity names have word overlap, and $0$ otherwise.

\subsection{Knowledge Storage}
\label{sec:storage}

While public KBs provide abundant world knowledge,  they are still unable to cover all knowledge that users are interested in. 
To meet users' personal knowledge demands,  
%Besides widely-used public KBs, 
% personalized KB, symbolic memory
KnowledGPT introduces a personalized KB (PKB) that acts as LLMs' symbolic memory, granting users the capability to store and access specialized knowledge. 
% extract
The PKB is populated by knowledge extracted from user-provided documents. 
When users intend to add knowledge into the PKB
we prompt LLMs to extract knowledge from their provided documents, with the prompt shown in Sec ~\ref{sec:prompts}.

We consider knowledge represented in three forms, including entity description, relational triples, and entity-aspect information, as is shown in Fig ~\ref{fig:challenges}, which is
different from RET-LLM~\citep{modarressi2023retllm} or the KG-Index of LangChain~\citep{Chase_LangChain_2022} and Llama Index~\citep{Liu_LlamaIndex_2022} that extract only triples,  
%We consider three types of knowledge, orienting entities in real world: entity description, relational triples, and entity aspect information. 
While entity description and relational triples have been widely adopted in knowledge bases like Wikipedia and Wikidata, they only represents a limited portion of knowledge, as is shown by experiments in Sec ~\ref{sec:pkb_exp}.
%enough for LLMs to use. 
For example, when we want to know the experience of Socrates as a soldier, most content in Socrates' wikipedia page would be hardly helpful, and it can also hardly be represented as a triple. 
Therefore, we propose an additional knowledge representation, termed as entity-aspect information, for symbolic memory of LLMs.  
It is a variation of triple where the object is a long piece of text which describes and can be retrieved by an entity and an aspect, 
%e.g. (\textit{``Socrates''}, \textit{``Military Service''}, \textit{``Socrates served as a Greek hoplite or heavy infantryman...''}).
%where the object is indexed by both an entity and an aspect, yielding detailed textual information. 
For instance, a record might be indexed by (\textit{``Socrates''}, \textit{``Military Service''}) and correspond to the description \textit{``Socrates served as a Greek hoplite or heavy infantryman...''}.
Knowledge represented in this form will also be retrieved also by the \textit{get\_entity\_or\_value} function. 

Given the tiny scale of PKBs in comparison to public KBs,  we consider a different strategy for entity linking on PKBs.
The difference is mainly three-fold.
First, we define the entity searching API for PKB based on exact match and embedding similarity. 
Embedding similarity aids in recognizing widely-known entity aliases, such as \textit{Chanelle Scott Calica} and \textit{Shystie}. 
Second, during extraction, an extracted entity mention is not aligned to entities existing in the PKB.  
Therefore, an entity may be extracted as distinct mentions in different documents.
%For example, the entity Bill Watts can be extracted as "Bill Watts" in one time and "William F. Watts Jr" in another time  , which are stored as two entities in the PKB. 
Hence, for entity linking, KnowledGPT returns multiple matched  entities. 
Third, 
an entity would be extracted with an aliases list, which would be provided to LLMs for entity linking.

For the \textit{get\_entity\_or\_value} function, since a relation can also be extracted as different expressions,  we opt to retrieve relations with similarity score over a threshold, instead of a top-scored relation.

\section{Experiments}

% experiment
In this section, we experiment KnowledGPT on various settings, 
including manually crafted diverse queries on popular KBs (Sec ~\ref{sec:query_kbs_exp}), knowledge-based question answering (Sec ~\ref{sec:kbqa}), and personalize KBs as memory for LLMs (Sec~\ref{sec:pkb_exp}).

\subsection{Experimental Setups}

\paragraph{\textbf{Knowledge Bases.}} KnowledGPT can access various KBs with its unified interface for KB operation. In this paper, we primarily consider the following KBs: 
\begin{enumerate}
    \item Wikipedia~\footnote{\url{https://en.wikipedia.org/}} and  Wikidata~\footnote{\url{https://wikidata.org/}}. Wikipedia provides rich encyclopedic information about world entities, maintained by global volunteers. 
    Wikidata is a knowledge graph complementing Wikipedia, which structures and organizes this encyclopedic knowledge in relational triples. 
    \item CN-DBPedia~\citep{xu2017cndbpedia} is a large-scale, constantly-updated Chinese KB extracted from various sources including Chinese Wikipedia~\footnote{\url{https://zh.wikipedia.org/}} and Baidu Baike~\footnote{\url{https://baike.baidu.com/}}. CN-DBPedia contains both entity descriptions like Wikipedia and relational triples like Wikidata. 
    \item Personalized KB is designed as a writable symbolic memory for LLMs. It stores upcoming knowledge extracted from user inputs.
    \item The NLPCC 2016 KBQA Knowledge Base~\citep{nlpcc2016}, which is widely adopted to evaluate models in terms of knowledge-based question answering tasks. 
    It contains 43 million triples. 
    This KB is only used in Sec ~\ref{sec:kbqa}.
    %43M SPO pairs with 6M subjects, 0.6M predicates and 16M objects
    %a Chinese KB provided specially for evaluating models performance on knowledge-based question answering tasks. We use it to evaluate KnowledGPT on knowledge-based question answering tasks.
\end{enumerate}

%In real applications, we will first detect the language of user query. If the user query is in Chinese, we use CN-DBPedia and Personalized KB. If the user query is in English, we use Wikidata and Personalized KB. 

In practical scenarios, we initiate the process by determining the language of the user query.
For English queries, Wikipedia and Wikidata, as well as  the personalized KB are employed.
For Chinese queries, we utilize CN-DBPedia in conjunction with the personalized KB. 
Our methods can also be easily extended to more languages with KBs in corresponding languages.

\paragraph{\textbf{Language Models.}} 
In this paper, we employ the powerful LLM \texttt{GPT-4} by default, which is accessed by the OpenAI API~\footnote{\url{https://platform.openai.com/}}. 
We prompt LLMs with instructions, requirements, and in-context examples, and require LLMs to output in json format. 
The detailed prompts are shown in Sec ~\ref{sec:prompts}. 
For sentence embeddings, we employ \texttt{GPT text-embedding-ada-002} model from OpenAI. 

\subsection{Queries on Popular KBs} 
\label{sec:query_kbs_exp}
\begin{table}[htbp]
    \small
    \centering
    \resizebox{\columnwidth}{!}{
    \begin{tabular}{ccccc}
    \Xhline{0.8pt}
        LLM & Direct Answer & Code & Entity Linking & KnowledGPT Answer \\ 
    \Xhline{0.8pt}
        %$\text{KnowledGPT}_\text{GPT-4}$ 
        GPT-4 & 4/11 & 9/11 & 22/22 & 11/11 \\
        %$\text{KnowledGPT}_\text{ChatGPT}$
        ChatGPT & 4/11 & 8/11 & 13/19 & 4/11 \\
        %Llama-2-Chat-13B & 0 & 0 & 0 & 0 \\
    \Xhline{0.8pt}
    \end{tabular}}
    \caption{Number of successful answers on the selected samples. The results are evaluated by human annotators. For code generation, it is considered successful if the code is supposed to be able to retrieve all the necessary knowledge related. }
    \label{tab:cases_acc}
\end{table} 

To evaluate KnowledGPT in terms of diversified queries of real users requiring external knowledge, we craft 11 questions about knowledge from CN-DBPedia~\footnote{As CN-DBPedia is a Chinese KB, the queries and generations are also in Chinese. We translate them into English to ease understanding. }. 
These questions span a variety of types, including single-hop and multi-hop relational queries, relation prediction, diversified instructions, mixed queries, and value comparison. 
These questions concern both popular entities and long-tail entities. 
The detailed examples are shown in Sec ~\ref{sec:detail_examples}.
We experiment on KnowledGPT with \texttt{GPT-4} and \texttt{ChatGPT} as the base LLMs, and measure the  rate of successful generation in terms of direct answers without KBs, code generation, entity linking, and KnowledGPT answers.

% 选择了11个样本，涉及
%To study the effectiveness of KnowledGPT towards various user queries requiring external knowledge, we manually construct 11 examples, concerning questions of the following types:
%single-hop relational search, relation prediction, diversified instruction, mixed query, multi-hop relational search, and value comparison. 
%\begin{enumerate}
    %\item Single-hop relational search, where the query contains an entity and a relation, and needs to find the target entity.
    %\item Relation prediction, where the query contains two entities, and models are expected to find their relationship based on KBs.
    %\item Diversified instruction, where the query not only requires knowledge, but also has diversified requirements.
    %\item Mixed query, where the query contains multiple sub queries that are even unrelated to each other. 
    %\item Multi-hop relational search, where the query can be decomposed into multiple sub queries, and models need to do multi-hop search on KBs to answer this query.
    %\item Value comparison, which not only requires multiple relational (attribute) search, but also needs to compare them. 
%\end{enumerate}

The results are shown in Table ~\ref{tab:cases_acc}. 
For detailed generations, please refer to Sec~\ref{sec:detail_examples}. 
We observe that:
(1) \texttt{GPT-4} and \texttt{ChatGPT} themselves are proficient at addressing queries about well-known entities, but they also hallucinate frequently about the unpopular entities. 
(2) KnowledGPT with \texttt{GPT-4} excellently accomplish tasks like code generation and entity linking, and eventually answers user queries with correct knowledge, representing a marked improvement over the vanilla responses from \texttt{GPT-4}.
(3) However, for \texttt{ChatGPT}, the success rate of intermediate steps remains to be improved, which constrains the overall efficacy of KnowledGPT. 
In the code generation step, \texttt{ChatGPT} sometimes generates poor relation aliases, such as \textit{who is the father},  especially for diverse or intricate queries.
% \texttt{ChatGPT} compared with GPT 4, relation aliases, bad, especially complex queries like multi-hop queries
This comparison suggests that \texttt{GPT-4} significantly outperforms \texttt{ChatGPT} in aspects like complex instruction understanding, task decomposition and code generation.
% Some examples contain entities that are both long-tail or popular
While we also experimented with smaller open-source LLMs like Llama-2-Chat-13B, they struggle to provide accurate answers directly and also fail to generate well-formatted code and respond in the JSON format as is required by the KnowledGPT framework.

% Complex Task, Json Parse fail 
% However, the results are not always in well json format, especially for smaller LLMs, like Llama-2-Chat-13B. 

%We further conduct case studies. 
% example multi hop reasoning
\paragraph{\textbf{Case Studies.}} 
Fig ~\ref{fig:full_example} illustrates a complete process of employing KnowledGPT to answer the question ``What are the titles of the poet who wrote \textit{Quiet Night Thoughts (Jing Ye Si)} ?'', which requires multi-hop knowledge retrieval and reasoning. 
% originally wrong
The original answer from \texttt{GPT-4} contains untruthful information, which shows the need to augment LLMs with precise external knowledge.
% code
KnowledGPT generates an excellent piece of code, which first looks for the author of the poem \textit{Quiet Night Thoughts}, and then searches for the author's titles. This demonstrates the effectiveness of solving multi-hop knowledge retrieval with code. 
% execute
%The code is successfully executed, which retrieves much relevant knowledge. 
Upon execution, the code efficiently retrieves relevant information.
% Noise
Finally, KnowledGPT integrates the retrieved knowledge to answer the query correctly. 
Despite potential noise in the intermediate steps, KnowledGPT well filters the noise and answers properly.

%\paragraph{\textbf{Entity Linking}}

\begin{figure}[tb]
    \centering
        \includegraphics[width=1\linewidth]{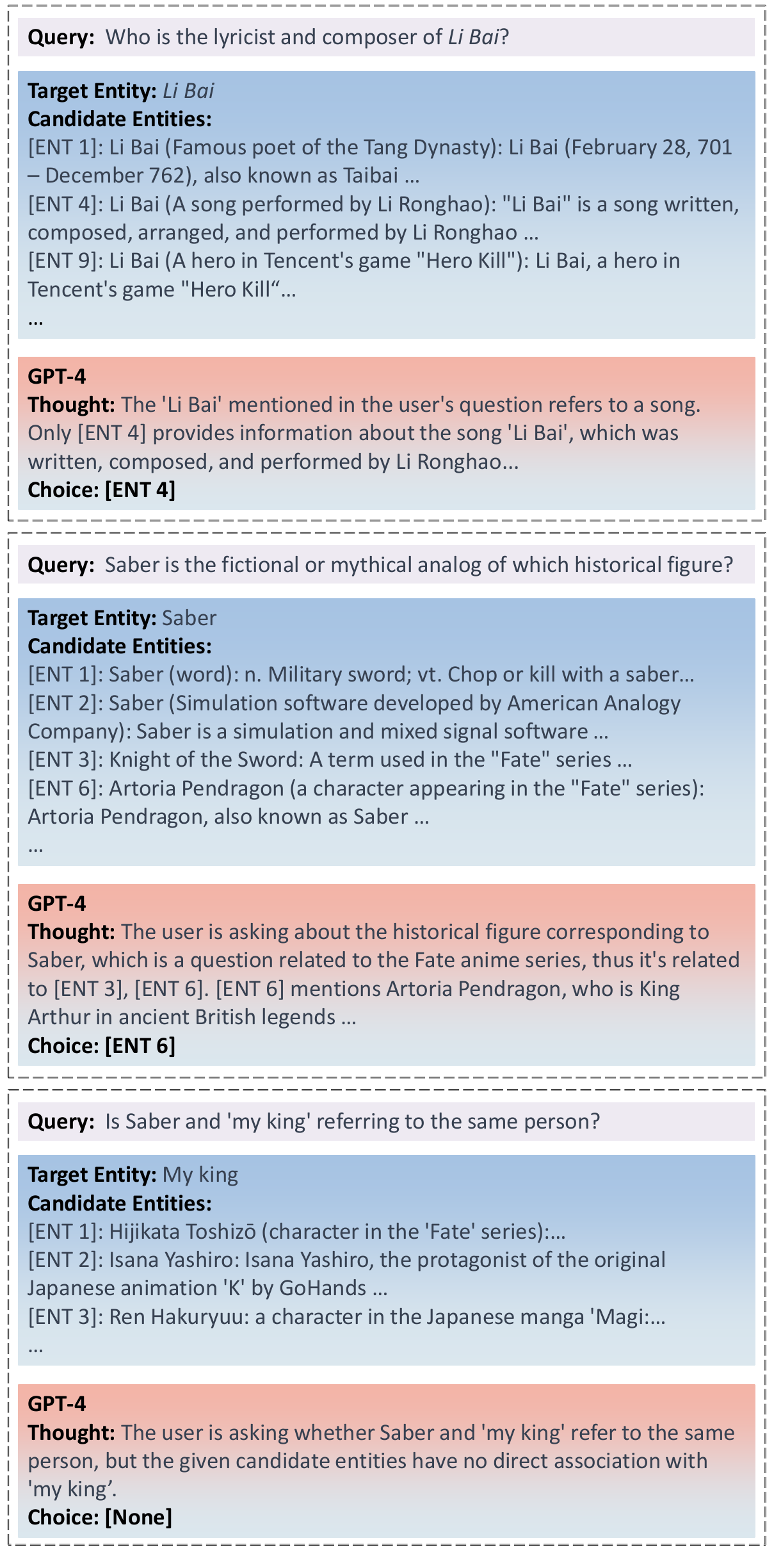} 
    \caption{Case studies of KnowledGPT's entity linking process. }
    \label{fig:el_example}
\end{figure} 

Fig ~\ref{fig:el_example} illustrates three instances of the entity linking step in KnowledGPT. 
% problem
The examples clearly indicate that the original candidate entities, returned from the entity linking and searching api of external KBs, are not well-ordered and might not even include the correct entity. 
Hence, simply selecting the top entity could introduce severe noise. 
% GPT-4
We apply \texttt{GPT-4} to select from the candidate entities provided with  their information. 
The results indicate that \texttt{GPT-4} proficiently identifies the correct entity for the query, and is also capable of rejecting all options when none is fit.

% GPT-4 Solution
%To mitigate this, we employ GPT-4 to handle the entity linking step, supplying it with the list of potential entities and their associated details.
%The outcomes illustrate that GPT-4 adeptly picks the right entity in relation to the query, and has the discernment to decline all candidates if none match the criteria.

\subsection{Knowledge-Based Question Answering}
~\label{sec:kbqa}
% KBQA
We evaluate KnowledGPT on zero-shot knowledge-based question answering (KBQA). 
KBQA is an extensively researched domain that seeks to answer natural language questions specific to certain relational triples in KBs. 
% Examlpe
For example, to answer the question ``Who is the author of \textit{the Republic}'', a KBQA model is expected to retrieve the triple (\textit{the Republic}, \textit{written by}, \textit{Plato}) and answer \textit{Plato}. 

% Our Dataset 
Given the expenses of invoking OpenAI APIs, we have compiled two compact datasets, namely NLPCC-100 for single-hop queries and NLPCC-MH-59 for multi-hop queries.   
NLPCC-100 is composed of 100 samples from the test set of NLPCC 2016 KBQA dataset~\citep{nlpcc2016}, 
while NLPCC-MH-59 consists of 59 samples from the test set of NLPCC-MH~\citep{wangyue2019}, a multi-hop KBQA dataset. 
NLPCC-MH is automatically constructed by expanding the NLPCC 2016 dataset, which leads to certain inherent noise.  
%We manually curated  59 samples from NLPCC-MH that are of high quality and have answer paths existing in the KB, and  corrected some noise.  
From the NLPCC-MH dataset, we manually select 59 samples, ensuring their quality and the presence of supporting fact chains in the KB. 
Furthermore, we rectify any noise present in these samples.
For both NLPCC-100 and NLPCC-MH-59, we use exclusively the full NLPCC 2016 KBQA Knowledge Base in this experiment.

%Hence, we select samples from NLPCC-MH with the following criteria: (1) answer paths exist in the kb, (2) the problem is logical and human is able to handle it, (3) remove the impurities in the answer and align the answer to the kb.

%yqw: NLPCC-MH is constructed by expanding the content of the NLPCC2016 one-hop questions to pay attention to Chinese multi-hop KBQA which can be answered by two-or-three paths in the NLPCC2016's knowledge base. 
% Strictly speaking, the quality of NLPCC-MH is average due to the way the dataset is constructed. So we selected two-hop questions from NLPCC-MH dataset with the following criteria: (1) answer paths exist in the kb, (2) the problem is logical and human is able to handle it, (3) remove the impurities in the answer and align the answer to the kb.

We make several modification to KnowledGPT towards this dataset and KB. 
First, in the provided KB, the tail entities of triples may contain multiple entities separated by special symbols, so we adjust the prompt for search code generation to request LLMs to include a splitting mechanism in the generated code.
% Our Process
%In the knowledge base provided, the objects of triples are more likely to contain multiple entities splited by different symbols, thus the code generated for knowledge retrieval is adjusted to split the results of the first round of query and perform the second round on the split entities in turn. 
Second, for better entity linking, as the provided KB does not contain entity description, we modify the implementation of \textit{\_get\_entity\_information} to return 10 triples related to the entity, sorted by the \texttt{jaccard} similarity between the query relation and the relation in triples. 
%\wxt{When nested in the \textit{find\_entity\_or\_value} function, it additionally considers relation aliases as input and prioritizes triples of similar relations for improved entity linking results.} For the \textit{get\_entity\_info} function, it outputs top-10 triples of the target entity based on the \texttt{jaccard} similarity of query and predicates of triples.
Third, we also modify the entity linking prompt, requiring LLMs to also adjust the query relation alias to better align with relations in the KB. 
%as for multi-hop QA, in the entity liking process, the LLM may skip the intermediate steps to modify the first round's relation aliases to the last round's relation aliases, thus it is required that new relation aliases generated in the entity linking process must be related to the given relation aliases.

% Baseline
We compare KnowledGPT with the following baseline methods :
\begin{enumerate}
    \item Retrieval via embedding similarity. Each triple is treated as a document and embedded using the CoSENT model~\citep{text2vec}.
    A single document is retrieved based on embedding similarity for each search. For multi-hop questions, the result from the first retrieval is added to the query to facilitate the second retrieval.
    %We consider two settings, namely the triple-level setting and the entity-level setting. In the former, we treat each triple as a document, which is the standard setting. In the later, we collect all triples of each entity as a document, and consider it correct if the document (entity) is retrieved, so this setting is simpler than the standard setting. We use CoSENT(Cosine Sentence) model~\citep{text2vec} for Chinese and Sentence-BERT model~\citep{reimers2019sentencebert} for English to embed the query and documents, calculate their cosine similarity, and retrieve the top document. For CoSENT model, we consider the standard setting for accuracy.
    \item Retrieval via BM25~\citep{robertson1995okapi}. 
    For each entity, we group all its triples as a document. 
    The most relevant document is retrieved using the BM25 algorithm for each search query, with stop words removed.
    We regard it as successful if the retrieved document contains the corresponding triple. 
    For multi-hop queries, we pick a triple from the initial retrieval's document based on the \texttt{jaccard} similarity between relations, and integrate this triple into the subsequent retrieval's query.
    %We only consider the entity-level setting for computational limits. We calculate the BM25 score between the query and documents, and retrieve the top document. 
    \item  SPE~\cite{lai2016open}. SPE extracts subject-predicate pairs from simple questions using embedding similarity. 
    %SPE obtains the first place in the contest of
    %used word vector similarity to score the candidates after entity linking and 
    It obtaied the first place in the contest of NLPCC 2016 KBQA task. 
\end{enumerate}

We report averaged F1 on NLPCC-100 and NLPCC-MH-59. 
% Introduce Averaged F1
Averaged F1 is a widely adopted metric for KBQA, designed for tasks with multiple golden answers and predictions. However, since in our dataset there is only one answer and one prediction for each sample, so the averaged F1 actually is equivalent to accuracy.

The results are shown in Table ~\ref{tab:kbqa1}, from which we have the following observation: 
(1) For single-hop queries, KnowledGPT significantly outperforms retrieval methods via BM25 and embedding similarity, which shows the effectiveness of retrieval from symbolic KBs compared with document corpus, for questions related to knowledge in KBs. 
(2) Zero-shot KnowledGPT outperforms the SPE method trained on the full training set of NLPCC-2016 KBQA dataset (0.92 vs 0.85), which shows the strong zero-shot performance of KnowledGPT. 
(3) For multi-hop queries, KnowledGPT also achieves excellent performance, while the performance of retrieval methods based on BM25 and embedding similarity drops significantly.

%For BM25, the averaged F1 score is 0.71 and most of the retrieved documents contain gold answers.
%The averaged F1 scores of questions and questions removing stop words~\footnote{We use baidu stop words in this paper.} are 0.31 and 0.28 respectively. 
%Based on Lai's results~\cite{lai2016open}, we select the answers of NLPCC2016-100 for which the averaged F1 score is 0.85.
%From these results, we have the following observations: 
%(1) better than other methods.
%(2) target relations for entity information retrival

%We compare KnowledGPT with ~\ref{tab:??} baselines on multi-hop questions NLPCC-MH. 
%With each question making multiple calls to the openai API and the high call spend, we chose a test subset of 59 from NLPCC-MH and the averaged F1 scores of these methods are shown in Table ~\ref{tab:kbqa2}.

\begin{table}[htbp]
    \small
    \centering
    \resizebox{\columnwidth}{!}{
    \begin{tabular}{ccccc}
    \Xhline{0.8pt}
        Dataset & BM25 & Embedding Similarity & SPE & KnowledGPT  \\ 
    
    \Xhline{0.8pt}
        NLPCC-100 & 0.71 & 0.31 & 0.85 & 0.92 \\
        NLPCC-MH-59 & 0.44 & 0.19 & - & 0.93 \\
    \Xhline{0.8pt}
    \end{tabular}}
    \caption{Averaged F1 of different methods on NLPCC-100 and NLPCC-MH-59.}
    \label{tab:kbqa1}
\end{table}

\subsection{KB as Memory}
\label{sec:pkb_exp}

\begin{figure}[tb]
    \centering
        \includegraphics[width=1\linewidth]{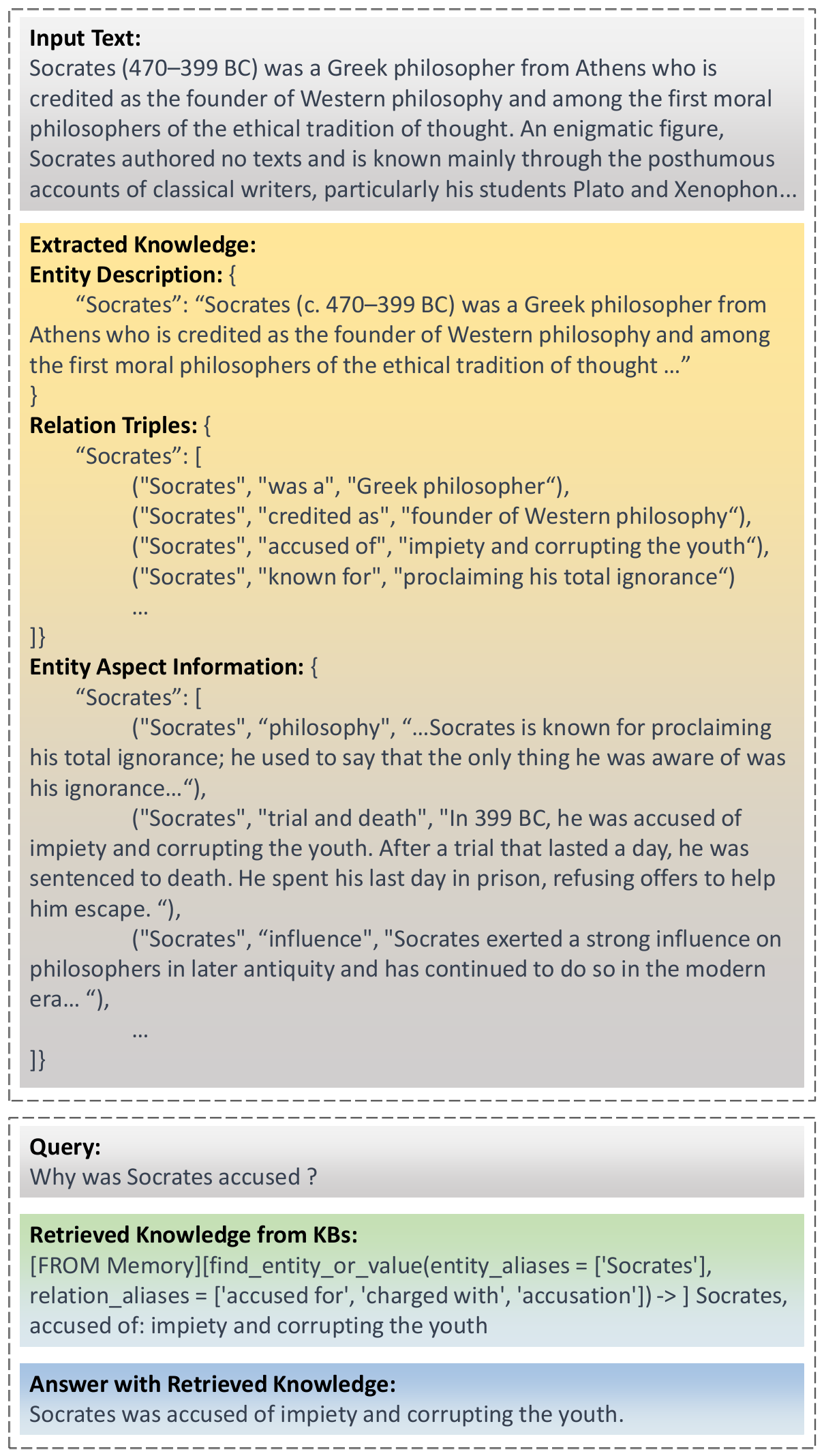} 
    \caption{An example showing knowledge extraction and retrieval of KnowledGPT on the personalized KB.}
    \label{fig:extracted_info}
\end{figure} 

We conduct experiments to study the effectiveness of KnowledGPT when paired with a modifiable personalized KB. 
KnowledGPT is tasked with building the PKB by extracting knowledge from provided documents, and study whether KnowledGPT answer corresponding questions properly with the PKB. 
Fig ~\ref{fig:extracted_info} shows an example
where text about Socrates is provided. KnowledGPT extracts knowledge from the text, and answers related questions by retrieving the extracted knowledge.

%Specifically, we base our experiments on the HotpotQA~\citep{yang-etal-2018-hotpotqa} dataset. 
We further apply KnowledGPT to HotpotQA~\citep{yang-etal-2018-hotpotqa}, a multi-hop question answering dataset with provided documents. 
We select 25 questions from HotpotQA dev set (distractor), including 5 comparison questions like ``Which is a shrub, Mimosa or Cryptocoryne?'' and 20 bridge questions like ``When was Erik Watts' father born?''. Each question is paired with 10 documents, and KnowledGPT extracts knowledge from these documents to build the PKB.

The results are shown in Table ~\ref{tab:hotpotqa}. 
KnowledGPT successfully answer all comparison questions, and 15 out of 20 bridge questions. 
Among the failed cases,  two bridge questions failed to extract the knowledge needed to answer the question and one bridge question failed in the entity linking stage.
Overall, this experiment suggests the promising future of utilizing PKBs as symbolic memory of LLMs.

%To further demonstrate the validity of KB as private memory of KnowledGPT, 25 questions are selected from HotpotQA dev set(distractor)~\cite{yang-etal-2018-hotpotqa}, a multihop question answering dataset. Every question is equipped with 10 documents from which knowledge extraction will be done. We select 5 comparison questions, like ``Which is a shrub, Mimosa or Cryptocoryne?'' and 20 bridge questions like ``When was Erik Watts' father born?''. The experiment results are shown in Table ~\ref{tab:hotpotqa}. Comparison questions are all correctly answered while 15 out of 20 bridge questions are correctly answered. Two bridge questions failed to extract the knowledge needed to answer the question and one bridge question failed in the entity linking stage.

\begin{table}[htbp]
    \small
    \centering
    \resizebox{\columnwidth}{!}{
    \begin{tabular}{cccc}
    \Xhline{0.8pt}
        & Comparison  & Bridge & All  \\ 
    \Xhline{0.8pt}
        KnowledGPT & 5/5 & 15/20 & 20/25 \\
    \Xhline{0.8pt}
    \end{tabular}}
    \caption{Number of successful answers on 25 questions selected from HotPotQA. The results are evaluated by human annotators. }
    \label{tab:hotpotqa}
\end{table}

We further investigate the knowledge extraction coverage of KnowledGPT on 100 documents from HotpotQA~\citep{yang-etal-2018-hotpotqa}, considering various LLMs and diverse combination of applied knowledge representations.
To quantify the coverage, we employ the word recall rate, calculated as
\begin{equation}
    \frac{ | W_{extracted} \cap W_{doc} |}{| W_{doc} |},
\end{equation}
where $|\cdot|$ indicates the cardinality of a set. 
$W_{extracted}$ and $W_{doc}$ denote the set of words in the extracted knowledge and the document respectively, after preprocessing including  the removal of stop words and lemmatization, utilizing the NLTK toolkit~\citep{nltk}.  
%$\frac{count(set(words\_extracted) \cap set(words\_in\_doc))}{count(set(words\_in\_doc))}$. 
%The words are preprocessed by removing stopping words and word lemmatization using the nltk toolset. 
%Knowledge extraction is performed using gpt3.5 and gpt4 for 100 documents sampled from hotpotqa dataset ~\cite{yang-etal-2018-hotpotqa}, and the types of knowledge extracted are entity description, relational triples, and entity-aspect-info. 
%In order to show the advantage of entity-aspect-info, both cases of extracting entity-aspect-info and not extracting entity-aspect-info are involved. 

The results are shown in Table ~\ref{tab:knowledge_coverage}, from which we have the following observations: 
(1) When restricting knowledge representation solely to triples, the extraction coverage stands at 0.53, which indicates that only a limited portion of knowledge can be represented as triples. Therefore, a PKB supporting triples alone falls short of adequately encompassing the knowledge provided by real users. 
(2) With additional  knowledge representations, i.e., entity description and entity-aspect information, we observe a marked improvement in knowledge extraction coverage, suggesting that incorporating entity description and entity-aspect information enables KnowledGPT to populate the PKB with a broader spectrum of knowledge.
(3) \texttt{ChatGPT} and \texttt{GPT-4} achieve similar proficiency for knowledge extraction. \texttt{GPT-4} outperforms \texttt{ChatGPT} only when entity-aspect info is included, which probably is attributed to \texttt{GPT-4}'s enhanced capability at following complex instructions. 

%Whether it's gpt3.5 or gpt4, the case of with entity-aspect-info is always better than the case of without entity-apect-info, proving the significance of entity-apect-info. 
%(2) In the case of without entity-aspect-info, gpt4 performs worse than gpt3.5. This may be because gpt4 is more compliant with the user requirement (of limiting the length of entity description) than gpt3.5, which sometimes still reproduces the original text in the entity description. 
%(3) In the case of with entity-aspect-info, gpt4 performs better than gpt3.5, showing that gpt4 is better at extracting entity-aspect-info than gpt3.5 using the current kg extraction prompt. 
%(4) In the case of extracting only triples, the results for gpt3.5 and gpt4 are the same.

\begin{table}[htbp]
    \small
    \centering
    \resizebox{\columnwidth}{!}{
    \begin{tabular}{cccc}
    \Xhline{0.8pt}
          %& without entity-aspect-info & with entity-aspect-info & with only triples \\ 
          & w/ triples only & + entity desc & ++ entity aspect info  \\ 
    
    \Xhline{0.8pt}
        %gpt3.5 & 0.66 & 0.81 & 0.53\\
        ChatGPT & 0.53 & 0.66 & 0.81\\
    \Xhline{0.8pt}
        %gpt4 & 0.62 & 0.86 & 0.53\\
        GPT-4 & 0.53 & 0.62 & 0.86\\
    \Xhline{0.8pt}
    \end{tabular}}
    \caption{Knowledge extraction coverage of KnowledGPT on 100 documents from HotpotQA with different LLMs and various combination of applied knowledge representations. }
    \label{tab:knowledge_coverage}
\end{table}

%According to our observation on various examples, KnowledGPT can extract a wide range of knowledge from the text with a satisfying coverage. However, there can also be omissions as well, especially for long documents.
% We found that LLMs are better at extracting knowledge at the beginning or end of the document, while they tend to ignore information in the middle.

\section{Limitations}
While KnowledGPT enables LLMs to effectively perform KB operations on external knowledge bases, 
there remain several limitations in its current form. 
First, the retrieval process entails a single-round of code generation and execution for efficiency concerns. 
However, a multi-round mechanism may better allow LLMs to autonomously explore KBs. 
As LLMs are not aware of the contents within KBs, they might generate search that appear logical but yield no results.
For example, a query like \textit{``Who is the voice actor for the heroine in ...'' } may require a two-hop searching for the relations \textit{heroine} and \textit{voice actor} subsequently in certain KBs, or just a single relation \textit{main voice actor} in others. 
In these scenarios, a multi-round mechanism empowers LLMs to probe and revisit the KBs autonomously, which might yield better results  but with increased costs.
Second, we experiment with KnowledGPT on representative yet small datasets, constrained by the expenses of accessing \texttt{GPT-4} via API. 
While the results validate the effectiveness of KnowledGPT, more comprehensive evaluations on full benchmarks are expected to better compare KnowledGPT to related methods. 
We plan to study fine-tuning LLMs like Llama for KnowledGPT in our future work to improve the efficiency and conduct more thorough experiments.
Finally, it remains a practical issue when LLMs need to access external KGs, rather than solving problems independently. 
In this work, we simply let LLMs make this decision, while better approaches remain to be explored.

%(1) Relation selection. 
%For example, what's the voice actor of the heroine in ...
%in some KBs, it requires two-hop reasoning.
%in other KBs, it 'main voice actor'. 

%No fail and retry, no multi-round. 
%LLMs don't know what's there in the KB, may generate search that sounds reasonale but retrieve nothing. 

%(2) Efficiency.
%multiple API, both openai api and KB api. 
%entity linking, require multiple interaction between LLM and KBs. 

\section{Conclusion}

In this paper, we introduce KnowledGPT, a comprehensive framework to integrate LLMs with external knowledge bases, facilitating LLMs' retrieval and storage on KBs.  
For retrieval, KnowledGPT adopts ``program of thought'' prompting, which  retrieves knowledge via code generation and execution.
For storage, KnowledGPT extracts various forms of knowledge from user provided texts, and populate the personalized KB with the extracted knowledge. 
KnowledGPT tackles several challenges inherent in  integrating LLMs with KBs, including complex question answering, ambiguation in entity linking, and limited forms of knowledge representations.
We show with extensive experiments that KnowledGPT effectively provides LLMs with the capability to operate on external KBs. 

%In the future, we will 
%(1) explore multi-round, autonomous KG agent.
%(2) decide when to use knowledge
%(3) evaluate fine-tuned LLMs for KnowledGPT on bigger datasets.

\bibliography{anthology,custom}

\begin{thebibliography}{45}
\expandafter\ifx\csname natexlab\endcsname\relax\def\natexlab#1{#1}\fi

\bibitem[{Abdelaziz et~al.(2021)Abdelaziz, Ravishankar, Kapanipathi, Roukos,
  and Gray}]{Abdelaziz_Ravishankar_Kapanipathi_Roukos_Gray_2021}
Ibrahim Abdelaziz, Srinivas Ravishankar, Pavan Kapanipathi, Salim Roukos, and
  Alexander Gray. 2021.
\newblock \href {https://doi.org/10.1609/aaai.v35i18.17988} {A semantic parsing
  and reasoning-based approach to knowledge base question answering}.
\newblock \emph{Proceedings of the AAAI Conference on Artificial Intelligence},
  35(18):15985--15987.

\bibitem[{Baek et~al.(2023)Baek, Aji, and Saffari}]{baek2023knowledge}
Jinheon Baek, Alham~Fikri Aji, and Amir Saffari. 2023.
\newblock Knowledge-augmented language model prompting for zero-shot knowledge
  graph question answering.
\newblock \emph{arXiv preprint arXiv:2306.04136}.

\bibitem[{Berant et~al.(2013)Berant, Chou, Frostig, and
  Liang}]{Berant2013SemanticPO}
Jonathan Berant, Andrew~K. Chou, Roy Frostig, and Percy Liang. 2013.
\newblock Semantic parsing on freebase from question-answer pairs.
\newblock In \emph{Conference on Empirical Methods in Natural Language
  Processing}.

\bibitem[{Bird and Klein(2009)}]{nltk}
Edward~Loper Bird, Steven and Ewan Klein. 2009.
\newblock Natural language processing with python.

\bibitem[{Brown et~al.(2020)Brown, Mann, Ryder, Subbiah, Kaplan, Dhariwal,
  Neelakantan, Shyam, Sastry, Askell, Agarwal, Herbert{-}Voss, Krueger,
  Henighan, Child, Ramesh, Ziegler, Wu, Winter, Hesse, Chen, Sigler, Litwin,
  Gray, Chess, Clark, Berner, McCandlish, Radford, Sutskever, and
  Amodei}]{gpt3}
Tom~B. Brown, Benjamin Mann, Nick Ryder, Melanie Subbiah, Jared Kaplan,
  Prafulla Dhariwal, Arvind Neelakantan, Pranav Shyam, Girish Sastry, Amanda
  Askell, Sandhini Agarwal, Ariel Herbert{-}Voss, Gretchen Krueger, Tom
  Henighan, Rewon Child, Aditya Ramesh, Daniel~M. Ziegler, Jeffrey Wu, Clemens
  Winter, Christopher Hesse, Mark Chen, Eric Sigler, Mateusz Litwin, Scott
  Gray, Benjamin Chess, Jack Clark, Christopher Berner, Sam McCandlish, Alec
  Radford, Ilya Sutskever, and Dario Amodei. 2020.
\newblock \href {http://arxiv.org/abs/2005.14165} {Language models are few-shot
  learners}.
\newblock \emph{CoRR}, abs/2005.14165.

\bibitem[{Cao et~al.(2022)Cao, Shi, Yao, Lv, Yu, Hou, Li, Liu, and
  Xiao}]{cao2022program}
Shulin Cao, Jiaxin Shi, Zijun Yao, Xin Lv, Jifan Yu, Lei Hou, Juanzi Li,
  Zhiyuan Liu, and Jinghui Xiao. 2022.
\newblock \href {http://arxiv.org/abs/2110.05743} {Program transfer for
  answering complex questions over knowledge bases}.

\bibitem[{Chase(2022)}]{Chase_LangChain_2022}
Harrison Chase. 2022.
\newblock \href {https://github.com/hwchase17/langchain} {{LangChain}}.

\bibitem[{Chen et~al.(2022)Chen, Ma, Wang, and Cohen}]{chen2022pot}
Wenhu Chen, Xueguang Ma, Xinyi Wang, and William~W Cohen. 2022.
\newblock Program of thoughts prompting: Disentangling computation from
  reasoning for numerical reasoning tasks.
\newblock \emph{arXiv preprint arXiv:2211.12588}.

\bibitem[{Chiang et~al.(2023)Chiang, Li, Lin, Sheng, Wu, Zhang, Zheng, Zhuang,
  Zhuang, Gonzalez, Stoica, and Xing}]{vicuna2023}
Wei-Lin Chiang, Zhuohan Li, Zi~Lin, Ying Sheng, Zhanghao Wu, Hao Zhang, Lianmin
  Zheng, Siyuan Zhuang, Yonghao Zhuang, Joseph~E. Gonzalez, Ion Stoica, and
  Eric~P. Xing. 2023.
\newblock \href {https://lmsys.org/blog/2023-03-30-vicuna/} {Vicuna: An
  open-source chatbot impressing gpt-4 with 90\%* chatgpt quality}.

\bibitem[{Duan(2016)}]{nlpcc2016}
Nan Duan. 2016.
\newblock Overview of the nlpcc-iccpol 2016 shared task: Open domain chinese
  question answering.
\newblock In \emph{Natural Language Understanding and Intelligent
  Applications}, pages 942--948. Springer International Publishing.

\bibitem[{Gu et~al.(2022)Gu, Deng, and Su}]{gu2022pangu}
Yu~Gu, Xiang Deng, and Yu~Su. 2022.
\newblock Don't generate, discriminate: A proposal for grounding language
  models to real-world environments.
\newblock \emph{arXiv preprint arXiv:2212.09736}.

\bibitem[{Guu et~al.(2020)Guu, Lee, Tung, Pasupat, and
  Chang}]{guu2020retrieval}
Kelvin Guu, Kenton Lee, Zora Tung, Panupong Pasupat, and Mingwei Chang. 2020.
\newblock Retrieval augmented language model pre-training.
\newblock In \emph{International conference on machine learning}, pages
  3929--3938. PMLR.

\bibitem[{Hao et~al.(2023)Hao, Liu, Wang, and Hu}]{hao2023toolkengpt}
Shibo Hao, Tianyang Liu, Zhen Wang, and Zhiting Hu. 2023.
\newblock Toolkengpt: Augmenting frozen language models with massive tools via
  tool embeddings.
\newblock \emph{arXiv preprint arXiv:2305.11554}.

\bibitem[{Hu et~al.(2023)Hu, Fu, Du, Luo, Zhao, and Zhao}]{hu2023chatdb}
Chenxu Hu, Jie Fu, Chenzhuang Du, Simian Luo, Junbo Zhao, and Hang Zhao. 2023.
\newblock Chatdb: Augmenting llms with databases as their symbolic memory.
\newblock \emph{arXiv preprint arXiv:2306.03901}.

\bibitem[{Jiang et~al.(2023)Jiang, Zhou, Zhao, and Wen}]{jiang2023unikgqa}
Jinhao Jiang, Kun Zhou, Wayne~Xin Zhao, and Ji-Rong Wen. 2023.
\newblock \href {http://arxiv.org/abs/2212.00959} {Unikgqa: Unified retrieval
  and reasoning for solving multi-hop question answering over knowledge graph}.

\bibitem[{Kandpal et~al.(2023)Kandpal, Deng, Roberts, Wallace, and
  Raffel}]{kandpal2023large}
Nikhil Kandpal, Haikang Deng, Adam Roberts, Eric Wallace, and Colin Raffel.
  2023.
\newblock Large language models struggle to learn long-tail knowledge.
\newblock In \emph{International Conference on Machine Learning}, pages
  15696--15707. PMLR.

\bibitem[{Kojima et~al.(2022)Kojima, Gu, Reid, Matsuo, and
  Iwasawa}]{kojima2022cot}
Takeshi Kojima, Shixiang~Shane Gu, Machel Reid, Yutaka Matsuo, and Yusuke
  Iwasawa. 2022.
\newblock Large language models are zero-shot reasoners.
\newblock \emph{Advances in neural information processing systems},
  35:22199--22213.

\bibitem[{Komeili et~al.(2022)Komeili, Shuster, and
  Weston}]{komeili-etal-2022-internet}
Mojtaba Komeili, Kurt Shuster, and Jason Weston. 2022.
\newblock \href {https://doi.org/10.18653/v1/2022.acl-long.579}
  {{I}nternet-augmented dialogue generation}.
\newblock In \emph{Proceedings of the 60th Annual Meeting of the Association
  for Computational Linguistics (Volume 1: Long Papers)}, pages 8460--8478,
  Dublin, Ireland. Association for Computational Linguistics.

\bibitem[{Lai et~al.(2016)Lai, Lin, Chen, Feng, and Zhao}]{lai2016open}
Yuxuan Lai, Yang Lin, Jiahao Chen, Yansong Feng, and Dongyan Zhao. 2016.
\newblock Open domain question answering system based on knowledge base.
\newblock In \emph{Natural Language Understanding and Intelligent Applications:
  5th CCF Conference on Natural Language Processing and Chinese Computing,
  NLPCC 2016, and 24th International Conference on Computer Processing of
  Oriental Languages, ICCPOL 2016, Kunming, China, December 2--6, 2016,
  Proceedings 24}, pages 722--733. Springer.

\bibitem[{Lazaridou et~al.(2022)Lazaridou, Gribovskaya, Stokowiec, and
  Grigorev}]{lazaridou2022internet}
Angeliki Lazaridou, Elena Gribovskaya, Wojciech Stokowiec, and Nikolai
  Grigorev. 2022.
\newblock Internet-augmented language models through few-shot prompting for
  open-domain question answering.
\newblock \emph{arXiv preprint arXiv:2203.05115}.

\bibitem[{Lewis et~al.(2020)Lewis, Perez, Piktus, Petroni, Karpukhin, Goyal,
  K{\"u}ttler, Lewis, Yih, Rockt{\"a}schel et~al.}]{lewis2020retrieval}
Patrick Lewis, Ethan Perez, Aleksandra Piktus, Fabio Petroni, Vladimir
  Karpukhin, Naman Goyal, Heinrich K{\"u}ttler, Mike Lewis, Wen-tau Yih, Tim
  Rockt{\"a}schel, et~al. 2020.
\newblock Retrieval-augmented generation for knowledge-intensive nlp tasks.
\newblock \emph{Advances in Neural Information Processing Systems},
  33:9459--9474.

\bibitem[{Liang et~al.(2016)Liang, Berant, Le, Forbus, and
  Lao}]{liang2016neural}
Chen Liang, Jonathan Berant, Quoc Le, Kenneth~D Forbus, and Ni~Lao. 2016.
\newblock Neural symbolic machines: Learning semantic parsers on freebase with
  weak supervision.
\newblock \emph{arXiv preprint arXiv:1611.00020}.

\bibitem[{Liu(2022)}]{Liu_LlamaIndex_2022}
Jerry Liu. 2022.
\newblock \href {https://doi.org/10.5281/zenodo.1234} {{LlamaIndex}}.

\bibitem[{Ming(2022)}]{text2vec}
Xu~Ming. 2022.
\newblock \href {https://github.com/shibing624/text2vec} {text2vec: A tool for
  text to vector}.

\bibitem[{Modarressi et~al.(2023)Modarressi, Imani, Fayyaz, and
  Sch{\"u}tze}]{modarressi2023retllm}
Ali Modarressi, Ayyoob Imani, Mohsen Fayyaz, and Hinrich Sch{\"u}tze. 2023.
\newblock Ret-llm: Towards a general read-write memory for large language
  models.
\newblock \emph{arXiv preprint arXiv:2305.14322}.

\bibitem[{OpenAI(2023)}]{openai2023gpt4}
OpenAI. 2023.
\newblock \href {http://arxiv.org/abs/2303.08774} {Gpt-4 technical report}.

\bibitem[{Prud'hommeaux(2011)}]{sparql}
Eric Prud'hommeaux. 2011.
\newblock \href {https://www.w3.org/TR/rdf-sparql-query/} {Sparql query
  language for rdf}.

\bibitem[{Robertson et~al.(1995)Robertson, Walker, Jones, Hancock-Beaulieu, and
  Gatford}]{robertson1995okapi}
Stephen Robertson, S.~Walker, S.~Jones, M.~M. Hancock-Beaulieu, and M.~Gatford.
  1995.
\newblock \href
  {https://www.microsoft.com/en-us/research/publication/okapi-at-trec-3/}
  {Okapi at trec-3}.
\newblock In \emph{Overview of the Third Text REtrieval Conference (TREC-3)},
  pages 109--126. Gaithersburg, MD: NIST.

\bibitem[{Schick et~al.(2023)Schick, Dwivedi-Yu, Dess{\`\i}, Raileanu, Lomeli,
  Zettlemoyer, Cancedda, and Scialom}]{schick2023toolformer}
Timo Schick, Jane Dwivedi-Yu, Roberto Dess{\`\i}, Roberta Raileanu, Maria
  Lomeli, Luke Zettlemoyer, Nicola Cancedda, and Thomas Scialom. 2023.
\newblock Toolformer: Language models can teach themselves to use tools.
\newblock \emph{arXiv preprint arXiv:2302.04761}.

\bibitem[{Shi et~al.(2021)Shi, Cao, Hou, Li, and Zhang}]{shi2021transfernet}
Jiaxin Shi, Shulin Cao, Lei Hou, Juanzi Li, and Hanwang Zhang. 2021.
\newblock \href {http://arxiv.org/abs/2104.07302} {Transfernet: An effective
  and transparent framework for multi-hop question answering over relation
  graph}.

\bibitem[{Sun et~al.(2019)Sun, Bedrax-Weiss, and Cohen}]{sun2019pullnet}
Haitian Sun, Tania Bedrax-Weiss, and William~W. Cohen. 2019.
\newblock \href {http://arxiv.org/abs/1904.09537} {Pullnet: Open domain
  question answering with iterative retrieval on knowledge bases and text}.

\bibitem[{Sun et~al.(2020)Sun, Zhang, Cheng, and Qu}]{sun2020sparqa}
Yawei Sun, Lingling Zhang, Gong Cheng, and Yuzhong Qu. 2020.
\newblock \href {http://arxiv.org/abs/2003.13956} {Sparqa: Skeleton-based
  semantic parsing for complex questions over knowledge bases}.

\bibitem[{Taori et~al.(2023)Taori, Gulrajani, Zhang, Dubois, Li, Guestrin,
  Liang, and Hashimoto}]{taori2023alpaca}
Rohan Taori, Ishaan Gulrajani, Tianyi Zhang, Yann Dubois, Xuechen Li, Carlos
  Guestrin, Percy Liang, and Tatsunori~B Hashimoto. 2023.
\newblock Alpaca: A strong, replicable instruction-following model.
\newblock \emph{Stanford Center for Research on Foundation Models.
  https://crfm. stanford. edu/2023/03/13/alpaca. html}, 3(6):7.

\bibitem[{Touvron et~al.(2023)Touvron, Lavril, Izacard, Martinet, Lachaux,
  Lacroix, Rozière, Goyal, Hambro, Azhar, Rodriguez, Joulin, Grave, and
  Lample}]{touvron2023llama}
Hugo Touvron, Thibaut Lavril, Gautier Izacard, Xavier Martinet, Marie-Anne
  Lachaux, Timothée Lacroix, Baptiste Rozière, Naman Goyal, Eric Hambro,
  Faisal Azhar, Aurelien Rodriguez, Armand Joulin, Edouard Grave, and Guillaume
  Lample. 2023.
\newblock \href {http://arxiv.org/abs/2302.13971} {Llama: Open and efficient
  foundation language models}.

\bibitem[{van Sonsbeek et~al.(2023)van Sonsbeek, Derakhshani, Najdenkoska,
  Snoek, and Worring}]{van2023open}
Tom van Sonsbeek, Mohammad~Mahdi Derakhshani, Ivona Najdenkoska, Cees~GM Snoek,
  and Marcel Worring. 2023.
\newblock Open-ended medical visual question answering through prefix tuning of
  language models.
\newblock \emph{arXiv preprint arXiv:2303.05977}.

\bibitem[{Wang et~al.(2023)Wang, Lyu, Ji, Zhang, Yu, Shi, and
  Tu}]{wang2023document}
Longyue Wang, Chenyang Lyu, Tianbo Ji, Zhirui Zhang, Dian Yu, Shuming Shi, and
  Zhaopeng Tu. 2023.
\newblock Document-level machine translation with large language models.
\newblock \emph{arXiv preprint arXiv:2304.02210}.

\bibitem[{Wang and Zhang(2019)}]{wangyue2019}
Yue Wang and Richong Zhang. 2019.
\newblock A dynamic programming-based approach to knowledge-basd question
  answering.
\newblock \emph{Journal of Zhengzhou University (Science Edition)},
  51(4):37--42.

\bibitem[{Xu et~al.(2017)Xu, Xu, Liang, Xie, Liang, Cui, and
  Xiao}]{xu2017cndbpedia}
Bo~Xu, Yong Xu, Jiaqing Liang, Chenhao Xie, Bin Liang, Wanyun Cui, and Yanghua
  Xiao. 2017.
\newblock Cn-dbpedia: A never-ending chinese knowledge extraction system.
\newblock In \emph{International Conference on Industrial, Engineering and
  Other Applications of Applied Intelligent Systems}, pages 428--438. Springer.

\bibitem[{Yang et~al.(2018)Yang, Qi, Zhang, Bengio, Cohen, Salakhutdinov, and
  Manning}]{yang-etal-2018-hotpotqa}
Zhilin Yang, Peng Qi, Saizheng Zhang, Yoshua Bengio, William Cohen, Ruslan
  Salakhutdinov, and Christopher~D. Manning. 2018.
\newblock \href {https://doi.org/10.18653/v1/D18-1259} {{H}otpot{QA}: A dataset
  for diverse, explainable multi-hop question answering}.
\newblock In \emph{Proceedings of the 2018 Conference on Empirical Methods in
  Natural Language Processing}, pages 2369--2380, Brussels, Belgium.
  Association for Computational Linguistics.

\bibitem[{Yao and Van~Durme(2014)}]{yao2014information}
Xuchen Yao and Benjamin Van~Durme. 2014.
\newblock Information extraction over structured data: Question answering with
  freebase.
\newblock In \emph{Proceedings of the 52nd annual meeting of the association
  for computational linguistics (volume 1: long papers)}, pages 956--966.

\bibitem[{Yu et~al.(2023)Yu, Zhang, Ng, Zhu, Li, Wang, Hu, Wang, Wang, and
  Xiang}]{yu2023decaf}
Donghan Yu, Sheng Zhang, Patrick Ng, Henghui Zhu, Alexander~Hanbo Li, Jun Wang,
  Yiqun Hu, William Wang, Zhiguo Wang, and Bing Xiang. 2023.
\newblock \href {http://arxiv.org/abs/2210.00063} {Decaf: Joint decoding of
  answers and logical forms for question answering over knowledge bases}.

\bibitem[{Zhang(2023)}]{zhang2023graphtoolformer}
Jiawei Zhang. 2023.
\newblock Graph-toolformer: To empower llms with graph reasoning ability via
  prompt augmented by chatgpt.
\newblock \emph{arXiv preprint arXiv:2304.11116}.

\bibitem[{Zhang et~al.(2022{\natexlab{a}})Zhang, Zhang, Yu, Tang, Tang, Li, and
  Chen}]{Zhang_2022}
Jing Zhang, Xiaokang Zhang, Jifan Yu, Jian Tang, Jie Tang, Cuiping Li, and Hong
  Chen. 2022{\natexlab{a}}.
\newblock \href {https://doi.org/10.18653/v1/2022.acl-long.396} {Subgraph
  retrieval enhanced model for multi-hop knowledge base question answering}.
\newblock In \emph{Proceedings of the 60th Annual Meeting of th Association for
  Computational Linguistics (Volume 1: Long Papers)}. Association for
  Computational Linguistics.

\bibitem[{Zhang et~al.(2022{\natexlab{b}})Zhang, Zhang, Li, and
  Zou}]{zhang2022crake}
Minhao Zhang, Ruoyu Zhang, Yanzeng Li, and Lei Zou. 2022{\natexlab{b}}.
\newblock \href {http://arxiv.org/abs/2207.03680} {Crake: Causal-enhanced
  table-filler for question answering over large scale knowledge base}.

\bibitem[{Zhang et~al.(2023)Zhang, Ladhak, Durmus, Liang, McKeown, and
  Hashimoto}]{zhang2023benchmarking}
Tianyi Zhang, Faisal Ladhak, Esin Durmus, Percy Liang, Kathleen McKeown, and
  Tatsunori~B Hashimoto. 2023.
\newblock Benchmarking large language models for news summarization.
\newblock \emph{arXiv preprint arXiv:2301.13848}.

\end{thebibliography}
\bibliographystyle{acl_natbib}

\appendix

\section{Prompts}
~\label{sec:prompts}

\begin{figure*}[htbp]
    \centering
        \includegraphics[width=1\linewidth]{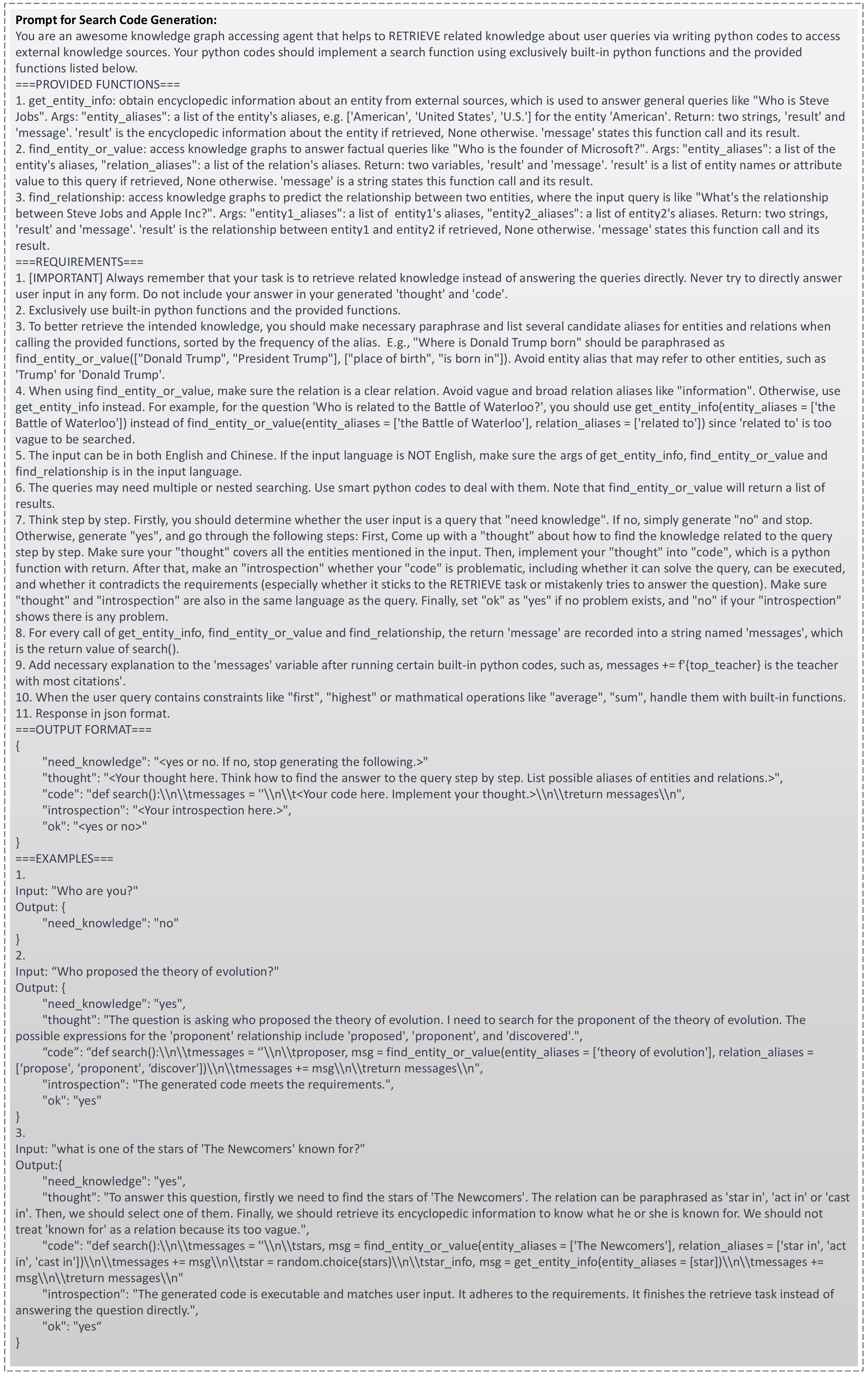} 
    \caption{Prompt for search code generation. The second example is in Chinese and translated into English. }
    \label{fig:prompt_search}
\end{figure*}

\begin{figure*}[htbp]
    \centering
        \includegraphics[width=1\linewidth]{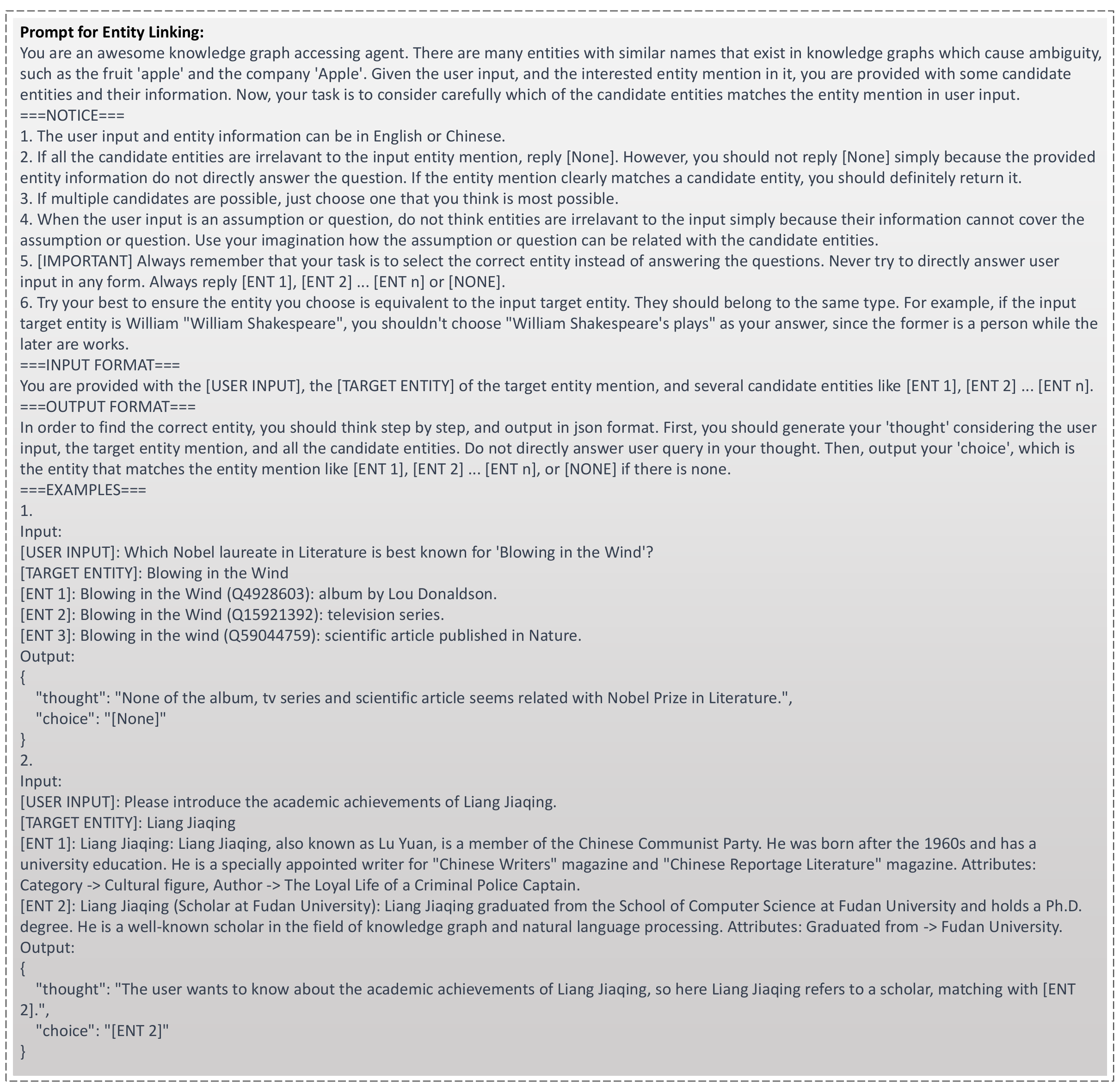} 
    \caption{Prompt for entity linking. The second example is in Chinese and translated into English.}
    \label{fig:prompt_el}
\end{figure*}

\begin{figure*}[htbp]
    \centering
        \includegraphics[width=1\linewidth]{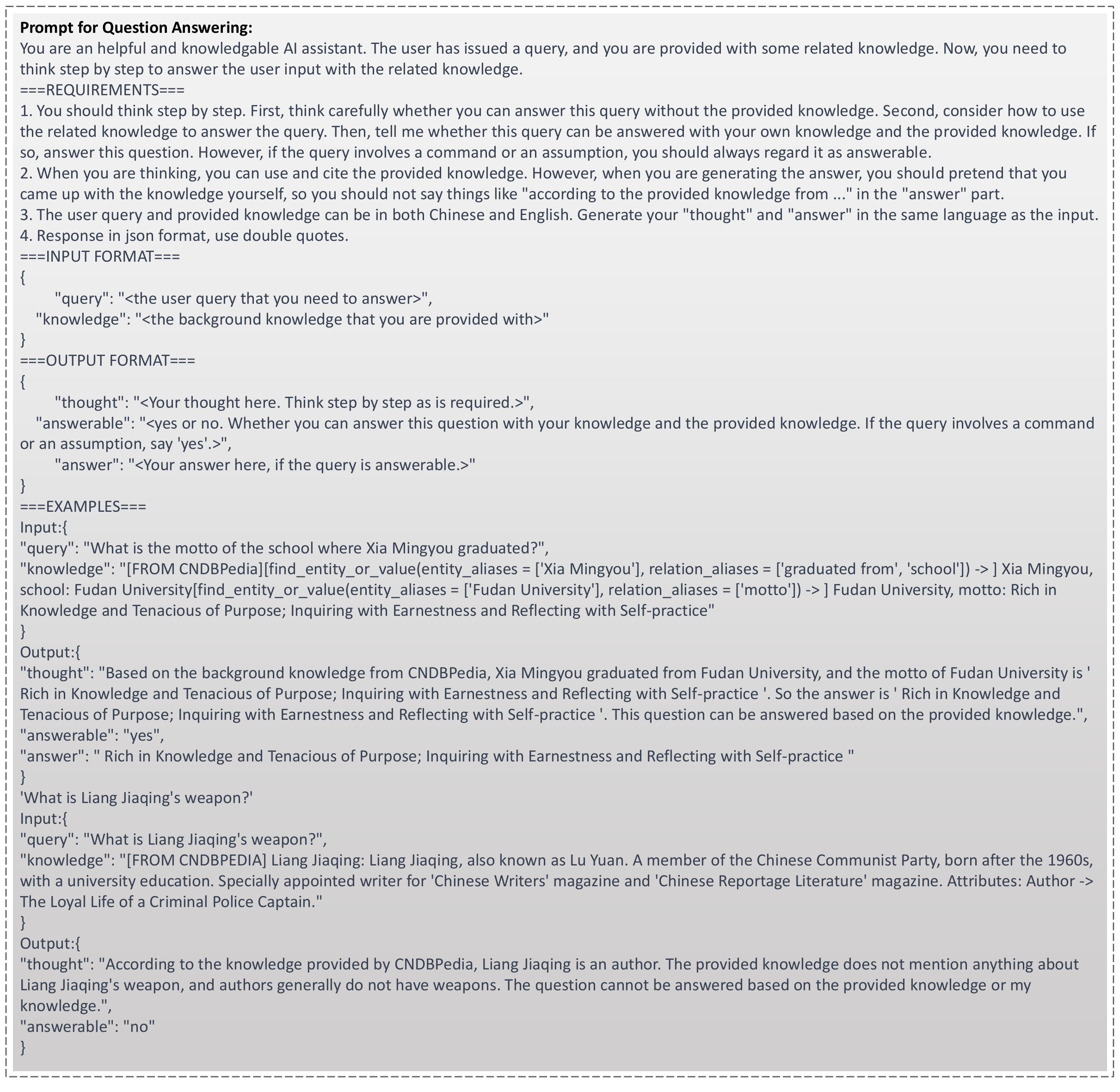} 
    \caption{Prompt for question answering. The examples are in Chinese and translated into English.}
    \label{fig:prompt_answer}
\end{figure*}

\begin{figure*}[htbp]
    \centering
        \includegraphics[width=1\linewidth]{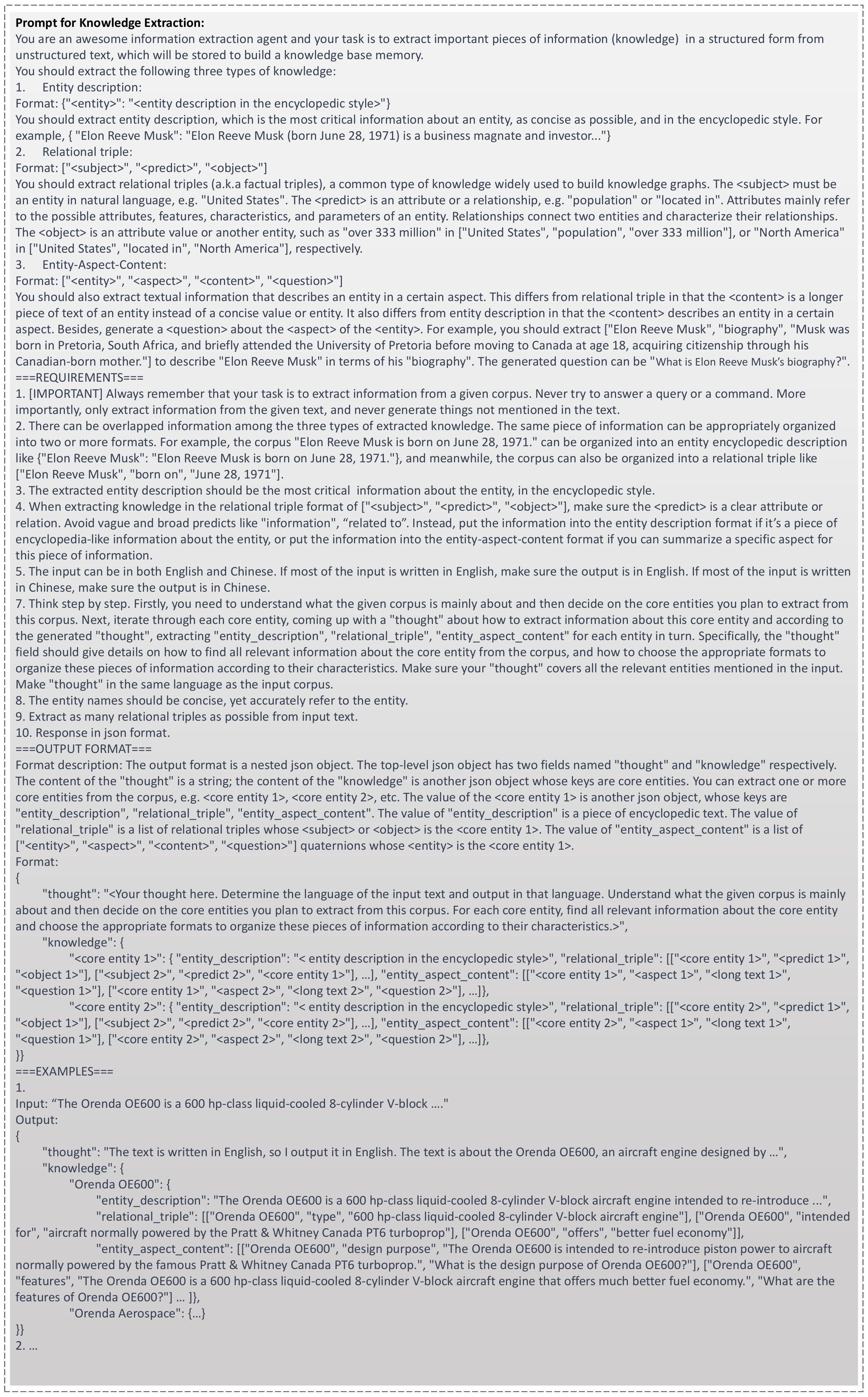} 
    \caption{Prompt for knowledge extraction. Certain contents about entity descriptions and in-context examples are omitted. }
    \label{fig:prompt_add}
\end{figure*}

The prompts are shown in Fig ~\ref{fig:prompt_search}, Fig ~\ref{fig:prompt_el}, Fig~\ref{fig:prompt_answer}, Fig~\ref{fig:prompt_add}. 
These prompts primarily consist of task introductions, requirements, the format of input and output, and in-context examples. 
We request LLMs to output in JSON format, and in the same language as the query. 
The prompts emphasize the LLMs should finish special tasks, instead of directly answering the input questions. 
They are mostly written in English, with some in-context examples in Chinese to better support queries in Chinese.
To ease understanding, these examples are translated into English in these figures. 

\paragraph{\textbf{Prompt for search code generation}} 
first asks LLMs to judge whether the query need knowledge from external KBs. 
If so, it continues to generate the search code. 
It pre-defines three functions that LLMs can generate to access KBs. It asks LLMs to generate a list of entity aliases or relation synonyms as input to these functions.

\paragraph{\textbf{Prompt for entity linking}}
directs LLMs to select from multiple candidate entities for the input query and target entity, provided with  information of the candidate entities.
While LLMs have the option to output [NONE] if none seems related, they are advised to use this option cautiously.
%LLMs can output [NONE] if none is believed to be appropriate, but we ask LLMs to output [NONE] cautiously.

\paragraph{\textbf{Prompt for question answering}} 
first guides LLMs to assess if the provided knowledge adequately supports  answering this question. 
If so, it asks LLMs to answer the query with the retrieved knowledge. 
Otherwise, we let LLMs answer this question independently.

\paragraph{\textbf{Prompt for knowledge extraction}} 
prompts LLMs to extract knowledge from the provided text. 
For long texts, LLMs tend to overlook many pieces of information during the  extraction proceess. 
We find that emphasizing  ``Do not miss any knowledge points.'' largely improves the knowledge extraction coverage. 
Additionally, the prompt encourages LLMs to present knowledge in the form of relational triples and entity-aspect information when possible, instead of entity description. 

%In order to make the extracted knowledge as complete as possible, the requirement of "Do not miss any knowledge points." is added in the prompt. It is worth noting that this sentence has a significant effect on gpt4, but not so much on gpt3.5. For short texts with less than 200 words, the extracted entity description tends to copy the original text, so it is emphasized in the prompt that the entity description can only be at most half the length of the original text, and gpt is told to use relational triples and the entity-aspect-info as much as possible to extract the knowledge. 

\section{Detailed Results of Queries on Popular KBs}
\label{sec:detail_examples}

In this section, we elaborate the detailed results of the manualy crafted examples mentioned in Sec ~\ref{sec:query_kbs_exp}. 
The examples are mainly in Chinese, including their queries, retrieval results, and answers. We translate them into English to ease understanding. 
We color incorrectly generated content that in red, 
including false facts or illogical code.
We also color unhelpful answers in brown, where \texttt{GPT-4} and \texttt{ChatGPT} admit their ignorance. 
In the two cases about value comparison, 
\texttt{GPT-4} attempts to conduct value comparison in the  code, which seems logical. 
However, it actually performs string comparison, which makes the results unreliable without improved value comparing functions that can take into account units and unit conversion.

\begin{table*}[htbp]
\sloppy
    \small
    \centering

    \resizebox{\textwidth}{!}{
    \begin{tabular}{|p{1cm}|p{7cm}|p{5cm}|p{5cm}|}
    \Xhline{0.8pt}
    % Exp 
    \multicolumn{4}{|l|}{\textbf{Input:} What is the registered capital of Dong Wu Securities?} \\ \hline 
        Model & Search Code & Retrieved Knowledge & Answer  \\ 
    \Xhline{0.8pt}
        % Model
        GPT-4  & 
        % Code
        def search(): \newline {\color{white}sp} messages = '' \newline {\color{white}sp} capital, msg = find\_entity\_or\_value(entity\_aliases = ['Dongwu Securities'], relation\_aliases = ['Registered Capital', 'Capital']) \newline {\color{white}sp} messages += msg \newline {\color{white}sp} return messages \newline
        & 
        % Knowledge
        [FROM CNDBPedia][find\_entity\_or\value(entity\_aliases = ['Dongwu Securities'], relation\_aliases = ['Registered Capital', 'Capital']) -> ] Dongwu Securities, Registered Capital: 1.5 billion Yuan &
        % Answer
        \textbf{W/ KB} \newline 
        1.5 billion Yuan. \newline  \newline 
        \textbf{W/o KB} \newline 
        \avoid{Sorry, I am unable to provide real-time specific data}, we suggest you visit the official website of Dong Wu Securities directly or contact their customer service for the most accurate information. \\
        \hline

        % Model
        ChatGPT  & 
        % Code
       def search(): \newline {\color{white}sp} messages = '' \newline {\color{white}sp} capital, msg = find\_entity\_or\_value(entity\_aliases = ['Dongwu Securities'], relation\_aliases = ['Registered Capital', 'Registered Capital']) \newline {\color{white}sp} messages += msg \newline {\color{white}sp} return messages \newline
       & 
        % Knowledge
        [FROM CNDBPedia][find\_entity\_or\value(entity\_aliases = ['Dongwu Securities'], relation\_aliases = ['Registered Capital', 'Registered Capital']) -> ] Dongwu Securities, Registered Capital: 1.5 billion Yuan

        &
        % Answer
        \textbf{W/ KB} \newline 
        1.5 billion Yuan. \newline  \newline 
        \textbf{W/o KB} \newline 
        \wrong{The registered capital of Soochow Securities is RMB 10 billion.}

        \\ \hline
    % Exp 
    \multicolumn{4}{|l|}{\textbf{Input:} Which historical hero is 
 Saber?} \\ \hline 
        Model & Search Code & Retrieved Knowledge & Answer  \\ 
    \Xhline{0.8pt}
        % Model
        GPT-4  & 
        % Code
        def search(): \newline {\color{white}sp} messages = '' \newline {\color{white}sp} historical\_figure, msg = find\_entity\_or\value(entity\_ aliases = ['Saber'], relation\_aliases = ['historical character', 'historical archetype']) \newline {\color{white}sp} messages += msg \newline {\color{white}sp} return messages \newline
        & 
        % Knowledge
        [FROM CNDBPedia][find\_entity\_or\_value(entity\_aliases = ['Saber'], relation\_aliases = ['Historical Character', 'Historical Archetype']) -> ] Artoria Pendragon (one of the female protagonists in the Japanese text-based adventure game "fate/ stay night"), Born: Ancient Britannia &
        % Answer
        \textbf{W/ KB} \newline 
        Artoria Pendragon. \newline  \newline 
        \textbf{W/o KB} \newline 
        Saber is a character from the Japanese anime Fate/stay night who is based on the legendary King Arthur.\\
        \hline

        % Model
        ChatGPT  & 
        % Code
       def search(): \newline {\color{white}sp} messages = '' \newline {\color{white}sp} hero, msg = \wrong{get\_entity\_info(entity\_aliases = ['Saber'], relation\_aliases = ['Historical heroes', 'Historical figures'])} \newline {\color{white}sp} messages += msg \newline {\color{white}sp} return messages \newline 
       & 
        % Knowledge
        None

        &
        % Answer
        \textbf{W/ KB} \newline 
        None \newline  \newline 
        \textbf{W/o KB} \newline 
        Saber is a character from the historical legend of King Arthur. In the legend of King Arthur, Saber is a brave and upright knight who possesses a legendary sword known as Excalibur.Saber has been featured in many works of literature, movies, and games, becoming a much loved and admired heroic figure.

        \\ \hline

    % Exp 
        \multicolumn{4}{|l|}{\textbf{Input:} Please write a resume for Prof. Sun Maosong that has a clear format and layout.} \\ \hline 
        Model & Search Code & Retrieved Knowledge & Answer  \\ 
    \Xhline{0.8pt}
        % Model
        GPT-4  & 
        % Code
        def search(): \newline {\color{white}sp} messages = '' \newline {\color{white}sp} info, msg = get\_entity\_info(entity\_aliases = ['Sun Maosong', 'Professor Sun Maosong']) ) \newline {\color{white}sp} messages += msg \newline {\color{white}sp} return messages \newline
        & 
        % Knowledge
        [FROM CNDBPedia][get\_entity\_info(entity\_aliases = ['Sun Maosong', 'Prof. Sun Maosong']) -> ] Sun Maosong (Secretary of the Party Committee of the Department of Computer Science, Tsinghua University): Sun Maosong, Professor, Doctoral Supervisor, was the head of the Department of Computer Science and Technology of Tsinghua University... Attributes: Category->Industry Figures, People. 
        &
        % Answer
        \textbf{W/ KB} \newline 
        Name: Sun Maosong
        
Position:

- Professor, Tsinghua University

- Doctoral Supervisor

...

Previous positions:

- Chairman, Department of Computer Science and Technology, Tsinghua University.
        \newline  \newline 
        \textbf{W/o KB} \newline 
         Biography of Prof. Maosong Sun
         
Personal Information:

...

Educational Experience:

1. \avoid{XX University, Ph.D., Major: XXXX, Year: XXXX-XXXX}
... \\
        \hline

        % Model
        ChatGPT  & 
        % Code
       def search(): \newline {\color{white}sp} messages = '' \newline {\color{white}sp} info, msg = get\_entity\_info(entity\_aliases = ['Prof. Sun Maosong']) \newline 
       {\color{white}sp} messages += msg \newline {\color{white}sp} return messages \newline
       & 
        % Knowledge
        [FROM CNDBPedia][get\_entity\_info(entity\_aliases = ['Prof. Sun Maosong']) -> ] Sun Maosong (Secretary of the Party Committee of the Department of Computer Science, Tsinghua University): Sun Maosong, Professor, Doctoral Supervisor, was the head of the Department of Computer Science and Technology of Tsinghua University... Attributes: Category->Industry Figures, People.  &
        % Answer
        \textbf{W/ KB} \newline 
        Sun Maosong
        
Position: Professor, Doctoral Supervisor

Work Experience:

- Chair, Department of Computer Science and Technology, Tsinghua University

...\newline  \newline 
        \textbf{W/o KB} \newline 
       Name: Sun Maosong
       
Educational Background:

- \avoid{Bachelor's degree: XX University, majoring in XXX, graduated in XXXX} ...

        \\ \hline
     
    \Xhline{0.8pt}
    \end{tabular}}
    \caption{ The detailed results of single-hop relational queries and a diversified instruction on CN-DBPedia.  }
    \label{tab:detail_1}
    
\end{table*} 

\begin{table*}[htbp]
\sloppy
    \small
    \centering

    \resizebox{\textwidth}{!}{
    \begin{tabular}{|p{1cm}|p{7cm}|p{5cm}|p{5cm}|}
    \Xhline{0.8pt}
    % Exp 
    \multicolumn{4}{|l|}{\textbf{Input:} What is the main ingredient in "Ants on a Tree"? Who is the voice of Saber?} \\ \hline 
        Model & Search Code & Retrieved Knowledge & Answer  \\ 
    \Xhline{0.8pt}
        % Model
        GPT-4  & 
        % Code
        def search(): \newline {\color{white}sp} messages = '' \newline {\color{white}sp} ingredients, msg = find\_entity\_or\_value(entity\_aliases = [ 'ants on a tree'], relation\_aliases = ['main ingredients', 'toppings']) \newline {\color{white}sp} messages += msg \newline {\color{white}sp} voice\_actor, msg = find\_ entity\_or\_value(entity\_aliases = ['Saber'], relation\_aliases = ['dub', 'voice actor']) \newline {\color{white}sp} messages += msg \newline {\color{white}sp} return messages \newline 
        & 
        % Knowledge
        [FROM CNDBPedia][find\_entity\_or\_value(entity\_aliases = ['Ants on a Tree'], relation\_aliases = ['Main Ingredients', 'Ingredients']) -> ] Ants on a Tree (Chinese name of the dish), Main Ingredients: Vermicelli (vermicelli), Minced Meat, Ginger, Garlic, Scallions, Onion [find\_entity\_or\_value(entity\_aliases = ['Saber'], relation\_aliases = ['Dubbing', 'Voice Actor']) -> ] saber (Spirit from the game "For Whom the Alchemy is Made"), Dubbed by: Ayako Kawasumi &
        % Answer
        \textbf{W/ KB} \newline 
        The main ingredient in Ants on a Tree is stir-fried vermicelli with minced pork, and Saber is voiced by Ayako Kawasumi. \newline  \newline 
        \textbf{W/o KB} \newline 
        The main ingredient of "Ants on a Tree" is stir-fried vermicelli with minced pork, and Saber is voiced by Ayako Kawasumi. \\
        \hline

        % Model
        ChatGPT  & 
        % Code
       def search(): \newline {\color{white}sp} messages = '' \newline {\color{white}sp} main\_ingredient, msg = find\_entity\_or\_value(entity\_ aliases = ['ants on a tree'], relation\_aliases = ['main ingredient', 'main ingredient', 'main ingredient']) \newline {\color{white}sp} messages += msg \newline {\color{white}sp} voice\_ actor, msg = find\_entity\_or\_value(entity\_aliases = ['Saber'], relation\_aliases = ['who the dub is', 'who the voice is', 'voice actor']) \newline {\color{white}sp} messages += msg \newline {\color{white}sp} return messages \newline 
       & 
        % Knowledge
        [FROM CNDBPedia][find\_entity\_or\_value(entity\_aliases = ['Ants on a tree'], relation\_aliases = ['Main Ingredients', 'Main Ingredients', 'Main Ingredients']) -> ] Ants on a tree (Chinese dish name), Main Ingredients: Vermicelli ( vermicelli), minced meat, ginger, garlic, scallions

        &
        % Answer
        \textbf{W/ KB} \newline 
        \avoid{No enough knowledge to answer.} \newline  \newline 
        \textbf{W/o KB} \newline 
        Ants on the Tree is a traditional Sichuan dish whose main ingredients are minced pork and vermicelli...
Regarding the voice of Saber, Saber is a character in the Fate/stay night series, and the voice actors for Saber vary depending on the production and version. Some of the more well-known voice actors are Ayako Kawasumi and Maya Sakamoto.

        \\ \hline

    % Exp 
        \multicolumn{4}{|l|}{\textbf{Input:} What are the titles of the poet writing Quiet Night Thoughts (Jing Ye Si) ?} \\ \hline 
        Model & Search Code & Retrieved Knowledge & Answer  \\ 
    \Xhline{0.8pt}
        % Model
        GPT-4  & 
        % Code
        def search(): \newline  {\color{white}sp} messages = '' \newline  {\color{white}sp} author, msg = find\_entity\_or\_value(entity\_aliases = ['Quiet Night Thoughts'], relation\_aliases = ['author', 'creator', 'writer']) \newline  {\color{white}sp} messages += msg \newline  {\color{white}sp} titles, msg = find\_entity\_or\_value(entity\_aliases = author, relation\_aliases = ['title', 'also known as', 'appellation']) \newline  {\color{white}sp} messages += msg \newline  {\color{white}sp} return messages \newline 
        & 
        % Knowledge
        [FROM CNDBPedia][find\_entity\_or\_value(entity\_aliases=['Quiet Night Thoughts'], relation\_aliases=['author', 'creator', 'writer'])->]"Quiet Night Thoughts" (A poem by Li Bai), Author: Li Bai, \wrong{The one who orders the destiny}.[find\_entity\_or\_value(entity\_aliases=['Li Bai','\wrong{The one who orders the destiny}'], relation\_aliases=['title', 'also known as', 'appellation'])->]Li Bai (Famous poet of the Tang Dynasty), also known as: Qinglian Jushi, Zhixianren. &
        % Answer
        \textbf{W/ KB} \newline 
        The poet of "Quiet Night Thoughts" has two titles: "Qinglian Jushi" and "Zhixianren". \newline  \newline 
        \textbf{W/o KB} \newline 
        The poet of "Quiet Night Thoughts" is Li Bai from the Tang Dynasty. He has several titles, including "Poetry Immortal", \wrong{"Wandering Poet"}, and "Wine Immortal". His poetic talent is outstanding, and his works are widely circulated, revered by later generations as the \wrong{"God of Poetry"}. \\
        \hline

        % Model
        ChatGPT  & 
        % Code
       def search(): \newline  {\color{white}sp} messages = '' \newline  {\color{white}sp} poet, msg = find\_entity\_or\_value(entity\_aliases = ['Quiet Night Thoughts'], relation\_aliases = \wrong{['author of "Quiet Night Thoughts"', 'creator of "Quiet Night
Thoughts"']}) \newline  {\color{white}sp} messages += msg \newline  {\color{white}sp} titles, msg = find\_entity\_or\_value(entity\_aliases = \wrong{[poet]}, relation\_aliases = ['title']) \newline  {\color{white}sp} messages += msg \newline  {\color{white}sp} return messages \newline  
       & 
        % Knowledge
        [FROM CNDBPedia][find\_entity\_or\_value(entity\_aliases = ['Quiet Night Thoughts'], relation\_aliases = ['author of "Quiet Night Thoughts"', 'creator of "Quiet Night
Thoughts"']) -> ] "Quiet Night Thoughts" (A poem by Li Bai), Author: Li Bai, \wrong{The one who orders the destiny}.

        &
        % Answer
        \textbf{W/ KB} \newline 
        \wrong{Li Bai}. \newline  \newline 
        \textbf{W/o KB} \newline 
        "Quiet Night Thoughts" is a work written by Tang Dynasty poet Li Bai. Li Bai is referred to as the "Poetry Immortal" and \wrong{"Poetry Sage"}, and he is also one of the most influential poets in Chinese literary history.

        \\ \hline

    % Exp 
        \multicolumn{4}{|l|}{\textbf{Input:} Do you know when Albert II's father was born?} \\ \hline 
        Model & Search Code & Retrieved Knowledge & Answer  \\ 
    \Xhline{0.8pt}
        % Model
        GPT-4  & 
        % Code
        def search(): \newline {\color{white}sp} messages = '' \newline {\color{white}sp} father, msg = find\_entity\_or\value(entity\_aliases = [' Albert II'], relation\_aliases = ['father', 'father is', 'dad']) \newline {\color{white}sp} messages += msg \newline {\color{white}sp} if father: \newline {\color{white}sp} {\color{white}sp} birth\_date, msg = find\_entity\_or\_value(entity\_aliases = father, relation\_aliases = ['birth\_date', 'born on']) \newline {\color{white}sp} {\color{white}sp} messages += msg \newline {\color{white}sp} return messages \newline
        & 
        % Knowledge
        [find\_entity\_or\_value(entity\_aliases = ['Albert II'], relation\_aliases = ['Father', 'Father was', 'Dad']) -> ] Albert II, Father: Rainier III [find\_entity\_or\_value(entity\_aliases = ['Albert II'], relation\_aliases = ['Father', 'Dad']) -> ] Albert II, Father: Rainier III [find\_entity\_or\_ value(entity\_aliases = ['Rainier III'], relation\_aliases = ['Date of Birth', 'Born']) -> ] Rainier III, Date of Birth: May 31, 1923&
        % Answer
        \textbf{W/ KB} \newline 
        May 31, 1923 .
        \newline  \newline 
        \textbf{W/o KB} \newline 
         Albert II's father was \wrong{Albert I}, who was born on \wrong{April 8, 1875}. \\
        \hline

        % Model
        ChatGPT  & 
        % Code
       def search(): \newline  {\color{white}sp} messages = '' \newline  {\color{white}sp} father, msg = find\_entity\_or\_value(entity\_aliases = ['Albert II'], relation\_aliases = \wrong{['who is the father', 'father of ']}) \newline  {\color{white}sp} messages += msg \newline  {\color{white}sp} if father: \newline  {\color{white}sp}  {\color{white}sp} birth\_date, msg = find\_entity\_or\_value(entity\_aliases = father, relation\_aliases = ['date of birth']) \newline  {\color{white}sp}  {\color{white}sp} messages += msg \newline  {\color{white}sp} return messages \newline
       & 
        % Knowledge
        None  &
        % Answer
        \textbf{W/ KB} \newline 
        None \newline  \newline 
        \textbf{W/o KB} \newline 
       Albert II's father was \wrong{Albert I}, who was born on \wrong{August 26, 1819}.

        \\ \hline
     
    \Xhline{0.8pt}
    \end{tabular}}
    \caption{ The detailed results of mixed queries and multi-hop relational  queries on CN-DBPedia.  }
    \label{tab:detail_1}
    
\end{table*} 

\begin{table*}[htbp]
\sloppy
    \small
    \centering

    \resizebox{\textwidth}{!}{
    \begin{tabular}{|p{1cm}|p{7cm}|p{5cm}|p{5cm}|}
    \Xhline{0.8pt}
    % Exp 
        \multicolumn{4}{|l|}{\textbf{Input:} Who were the  winners of the first Nobel Prize? What prizes did they each receive?} \\ \hline 
        Model & Search Code & Retrieved Knowledge & Answer  \\ 
    \Xhline{0.8pt}
        % Model
        GPT-4  & 
        % Code
        def search(): \newline  {\color{white}sp} messages = '' \newline  {\color{white}sp} first\_winners, msg = find\_entity\_or\_value(entity\_aliases = ['Nobel Prize'], relation\_aliases = ['first winner', 'first recipient']) \newline  {\color{white}sp} messages += msg \newline  {\color{white}sp} for winner in first\_winners: \newline  {\color{white}sp}  {\color{white}sp} award, msg = find\_entity\_or\_value(entity\_aliases = [winner], relation\_aliases = ['awarded', 'award']) \newline  {\color{white}sp}  {\color{white}sp} messages += msg \newline  {\color{white}sp} return messages \newline 
        & [FROM CNDBPedia][find\_entity\_or\_value(entity\_aliases = ['Nobel Prize'], relation\_aliases = ['first winner', 'first recipient']) -> ] Nobel Prize, first winner: Wilhelm Conrad Röntgen, Jacobus Henricus van 't Hoff, Emil Adolf von Behring, Sully Prudhomme, Henry Dunant, Frédéric Passy.[find\_entity\_or\_value(entity\_aliases = ['Wilhelm Conrad Röntgen'], relation\_aliases = ['awarded', 'award']) -> ] Wilhelm Conrad Röntgen, \wrong{Occupation}: Physicist.
        ...
        [find\_entity\_or\_value(entity\_aliases = ['Frédéric Passy'], relation\_aliases = ['awarded', 'award']) -> ] Frédéric Passy, Major Achievement: First recipient of the Nobel Peace Prize. &
        % Answer
        \textbf{W/ KB} \newline 
         The inaugural winners of the Nobel Prize included Wilhelm Conrad Röntgen, Jacobus Henricus van 't Hoff, Emil Adolf von Behring, Sully Prudhomme, Henry Dunant, and Frédéric Passy. They respectively received the Nobel Prizes in Physics, Chemistry, Medicine, Literature, and Peace. \newline  \newline 
        \textbf{W/o KB} \newline 
        The inaugural winners of the Nobel Prize were:
1. Nobel Prize in Physics: Wilhelm Conrad Röntgen, who ...
2. Nobel Prize in Chemistry: Jacobus Henricus van 't Hoff, who ...
3. Nobel Prize in Physiology or Medicine: Emil Adolf von Behring, who ...
4. Nobel Prize in Literature: Sully Prudhomme, who ...
5. Nobel Peace Prize: Henry Dunant, who ... Frédéric Passy, who ... \\
        \hline

        % Model
        ChatGPT  & 
        % Code
       def search(): \newline  {\color{white}sp} messages = '' \newline  {\color{white}sp} first\_winners, msg = find\_entity\_or\_value(entity\_aliases = ['Nobel Prize'], relation\_aliases = ['first winner', 'first recipient']) \newline  {\color{white}sp} messages += msg \newline  {\color{white}sp} awards = [] \newline  {\color{white}sp} for winner in first\_winners: \newline  {\color{white}sp}  {\color{white}sp} award, msg = find\_entity\_or\_value(entity\_aliases = [winner], relation\_aliases = ['award']) \newline  {\color{white}sp}  {\color{white}sp} messages += msg \newline  {\color{white}sp}  {\color{white}sp} awards.append(award[0]) \newline  {\color{white}sp} messages += f'first winner is \{first\_winners\}, they respectively received \{awards\}.' \newline  {\color{white}sp} return messages \newline  
       & 
        % Knowledge
        [FROM CNDBPedia][find\_entity\_or\_value(entity\_aliases = ['Nobel Prize'], relation\_aliases = ['first winner', 'first recipient']) -> ] Nobel Prize, first winner: Wilhelm Conrad Röntgen, Jacobus Henricus van 't Hoff, Emil Adolf von Behring, Sully Prudhomme, Henry Dunant, Frédéric Passy.[find\_entity\_or\_value(entity\_aliases = ['Wilhelm Conrad Röntgen'], relation\_aliases = ['award']) -> ] Wilhelm Conrad Röntgen, Occupation: Physicist.
        ...
        \wrong{(Did not retrieve results for all the  laureates. )}
        &
        % Answer
        \textbf{W/ KB} \newline 
        Wilhelm Conrad Röntgen received the Physics Prize, Jacobus Henricus van 't Hoff received the Chemistry Prize, and Emil Adolf von Behring received the Medicine Prize. \wrong{(Miss three winners.)} \newline  \newline 
        \textbf{W/o KB} \newline 
        The first Nobel Prizes were awarded in 1901. Here are the inaugural Nobel Prize winners and their respective categories:
1. Physics Prize: Wilhelm Conrad Röntgen and \wrong{Wilhelm Hertz}
2. Chemistry Prize: Jacobus Hansen van 't Hoff and \wrong{Hermann Emil Fischer}
3. Physiology or Medicine Prize: Emil von Behring and \wrong{William Ramsay}
4. Literature Prize: \wrong{José Echegaray and Selma Lagerlöf}
5. Peace Prize: Henry Dunant and Frédéric Passy
...

        \\ \hline

    % Exp 
        \multicolumn{4}{|l|}{\textbf{Input:} Is the lyricist, composer, and arranger of \textit{Li Bai} the same person?} \\ \hline 
        Model & Search Code & Retrieved Knowledge & Answer  \\ 
    \Xhline{0.8pt}
        % Model
        GPT-4  & 
        % Code
        def search(): \newline  {\color{white}sp} messages = '' \newline  {\color{white}sp} lyricist, msg = find\_entity\_or\_value(entity\_aliases = ['Li Bai'], relation\_aliases = ['lyrics writing', 'lyricist']) \newline  {\color{white}sp} messages += msg \newline  {\color{white}sp} composer, msg = find\_entity\_or\_value(entity\_aliases = ['Li Bai'], relation\_aliases = ['compose', 'composer']) \newline  {\color{white}sp} messages += msg \newline  {\color{white}sp} arranger, msg = find\_entity\_or\_value(entity\_aliases = ['Li Bai'], relation\_aliases = ['arrange', 'arranger']) \newline  {\color{white}sp} messages += msg \newline  {\color{white}sp} if lyricist == composer == arranger: \newline  {\color{white}sp}  {\color{white}sp} messages += 'The lyricist, composer, and arranger of ''Li Bai'' are the same person.' \newline  {\color{white}sp} else: \newline  {\color{white}sp}  {\color{white}sp} messages += 'The lyricist, composer, and arranger of ''Li Bai'' are not the same person.' \newline  {\color{white}sp} return messages \newline  & 
        % Knowledge
        [FROM CNDBPedia][find\_entity\_or\_value(entity\_aliases = ['Li Bai'], relation\_aliases = ['lyrics writing', 'lyricist']) -> ] Li Bai (a song sung by Li Ronghao), lyricist: Li Ronghao[find\_entity\_or\_value(entity\_aliases = ['Li Bai'], relation\_aliases = ['compose', 'composer']) -> ] Li Bai (a song sung by Li Ronghao), \wrong{arrange}: Li Ronghao [find\_entity\_or\_value(entity\_aliases = ['Li Bai'], relation\_aliases = ['arrange', 'arranger']) -> ] Li Bai (a song sung by Li Ronghao), arrange: Li Ronghao. The song \textit{Li Bai} has its lyrics, composition, and arrangement all done by the same person. &
        % Answer
        \textbf{W/ KB} \newline 
        Yes, the song \textit{Li Bai} has its lyrics, composition, and arrangement all done by the same person, Li Ronghao. \newline  \newline 
        \textbf{W/o KB} \newline 
         The lyricist, composer, and arranger of the song \textit{Li Bai} are \wrong{not} the same person. The lyrics and composition of this song were done \wrong{by Ashin (A Xin/A Pa)}, while the arrangement was done \wrong{jointly by Mayday and Goro Liu.}  \\
        \hline

        % Model
        ChatGPT  & 
        % Code
       def search(): \newline  {\color{white}sp} messages = '' \newline  {\color{white}sp} lyricist, msg = find\_entity\_or\_value(entity\_aliases = ['Li Bai'], relation\_aliases = ['lyrics writing']) \newline  {\color{white}sp} messages += msg \newline  {\color{white}sp} composer, msg = find\_entity\_or\_value(entity\_aliases = ['Li Bai'], relation\_aliases = ['compose']) \newline  {\color{white}sp} messages += msg \newline  {\color{white}sp} arranger, msg = find\_entity\_or\_value(entity\_aliases = ['Li Bai'], relation\_aliases = ['arrange']) \newline  {\color{white}sp} messages += msg \newline  {\color{white}sp} if len(lyricist) == 1 and len(composer) == 1 and len(arranger) == 1 and lyricist[0] == composer[0] == arranger[0]: \newline  {\color{white}sp}  {\color{white}sp} messages += 'Is the same person' \newline  {\color{white}sp} else: \newline  {\color{white}sp}  {\color{white}sp} messages += 'Not the same person' \newline  {\color{white}sp} return messages \newline 
       & 
        % Knowledge
        None &
        % Answer
        \textbf{W/ KB} \newline 
        None \newline  \newline 
        \textbf{W/o KB} \newline 
       The lyrics, composition, and arrangement of \textit{Li Bai} were all done by the same person, the composer Li Ronghao.

        \\ \hline
     
    \Xhline{0.8pt}
    \end{tabular}}
    \caption{ The detailed results of multi-hop relational queries on CN-DBPedia.  }
    \label{tab:detail_1}
    
\end{table*}

\begin{table*}[htbp]
\sloppy
    \small
    \centering

    \resizebox{\textwidth}{!}{
    \begin{tabular}{|p{1cm}|p{7cm}|p{5cm}|p{5cm}|}
    \Xhline{0.8pt}
    % Exp 
        \multicolumn{4}{|l|}{\textbf{Input:} What is the relationship between Ronald Lee and Li Bai?} \\ \hline 
        Model & Search Code & Retrieved Knowledge & Answer  \\ 
    \Xhline{0.8pt}
        % Model
        GPT-4  & 
        % Code
        def search(): \newline  {\color{white}sp} messages = '' \newline  {\color{white}sp} relationship, msg = find\_relationship(entity1\_aliases = ['Li Ronghao'], entity2\_aliases = ['Li Bai']) \newline  {\color{white}sp} messages += msg \newline  {\color{white}sp} return messages \newline 
        & 
        % Knowledge
        [FROM CNDBPedia][find\_relationship(entity1\_aliases = ['Li Ronghao'], entity2\_aliases = ['Li Bai']) -> ] Li Ronghao, Representative Work: Li Bai &
        % Answer
        \textbf{W/ KB} \newline 
        Li Ronghao created a song titled 'Li Bai' which is about his relationship with Li Bai. \newline  \newline 
        \textbf{W/o KB} \newline 
        \avoid{Li Ronghao and Li Bai are not directly related.} Li Ronghao is a famous singer and music producer in modern times... \\
        \hline

        % Model
        ChatGPT  & 
        % Code
       def search(): \newline  {\color{white}sp} messages = '' \newline  {\color{white}sp} relationship, msg = find\_relationship(entity1\_aliases = ['Li Ronghao'], entity2\_aliases = ['Li Bai']) \newline  {\color{white}sp} messages += msg \newline  {\color{white}sp} return messages \newline 
       & 
        % Knowledge
        [FROM CNDBPedia][find\_relationship(entity1\_aliases = ['Li Ronghao'], entity2\_aliases = ['Li Bai']) -> ] Li Ronghao, Representative Work: Li Bai
        &
        % Answer
        \textbf{W/ KB} \newline 
        \wrong{Li Ronghao and Li Bai are the same person}. \newline  \newline 
        \textbf{W/o KB} \newline 
       \avoid{Li Ronghao and Li Bai are two different people with no direct relationship.} Li Ronghao is a famous singer and music producer in modern China, while Li Bai is ...

        \\ \hline
    % Exp 
        \multicolumn{4}{|l|}{\textbf{Input:} Who is taller, Yao Ming or Saber?} \\ \hline 
        Model & Search Code & Retrieved Knowledge & Answer  \\ 
    \Xhline{0.8pt}
        % Model
        GPT-4  & 
        % Code
        def search(): \newline  {\color{white}sp} messages = '' \newline  {\color{white}sp} yao\_height, msg = find\_entity\_or\_value(entity\_aliases = ['Yao Ming'], relation\_aliases = ['height']) \newline  {\color{white}sp} messages += msg \newline  {\color{white}sp} saber\_height, msg = find\_entity\_or\_value(entity\_aliases = ['Saber'], relation\_aliases = ['height']) \newline  {\color{white}sp} messages += msg \newline  {\color{white}sp} if \wrong{yao\_height[0] > saber\_height[0]}: \newline  {\color{white}sp}  {\color{white}sp} messages += 'Yao Ming is taller than Saber.' \newline  {\color{white}sp} elif \wrong{yao\_height[0] < saber\_height[0]}: \newline  {\color{white}sp}  {\color{white}sp} messages += 'Saber is taller than Yao Ming' \newline  {\color{white}sp} else: \newline  {\color{white}sp}  {\color{white}sp} messages += 'Yao Ming and Saber are of the same height.' \newline  {\color{white}sp} return messages & 
        % Knowledge
        [FROM CNDBPedia][find\_entity\_or\_value(entity\_aliases = ['Yao Ming'], relation\_aliases = ['height', 'altitude']) -> ] Yao Ming (Chairman of the Asian Basketball Association, Chairman of the Chinese Basketball Association), height: 226 cm [find\_entity\_or\_value(entity\_aliases = ['Saber'], relation\_aliases = ['height', 'altitude']) -> ] saber , height: 154 cm Yao Ming is taller than Saber." &
        % Answer
        \textbf{W/ KB} \newline 
        Yao Ming is taller than Saber. \newline  \newline 
        \textbf{W/o KB} \newline 
        Yao Ming's height is 2.26 meters, and Saber (a fictional character from the Fate series) is 1.54 meters tall. So, Yao Ming is taller than Saber. \\
        \hline

        % Model
        ChatGPT  & 
        % Code
       def search(): \newline  {\color{white}sp} messages = '' \newline  {\color{white}sp} yao\_height, msg = find\_entity\_or\_value(entity\_aliases = ['Yao Ming'], relation\_aliases = ['height']) \newline  {\color{white}sp} messages += msg \newline  {\color{white}sp} saber\_height, msg = find\_entity\_or\_value(entity\_aliases = ['Saber'], relation\_aliases = ['height']) \newline  {\color{white}sp} messages += msg \newline  {\color{white}sp} return messages \newline  
       & 
        % Knowledge
        [FROM CNDBPedia][find\_entity\_or\_value(entity\_aliases = ['Yao Ming'], relation\_aliases = ['height']) -> ] Yao Ming (Chairman of the Asian Basketball Association, Chairman of the Chinese Basketball Association), height: 226 cm 
        &
        % Answer
        \textbf{W/ KB} \newline 
        Yao Ming is taller than Saber. \newline  \newline 
        \textbf{W/o KB} \newline 
        Yao Ming and Saber are figures from completely different fields, making it difficult to directly compare their heights. Yao Ming is a famous basketball player from China, standing at 2.26 meters, while Saber is a character from the Japanese animation "Fate/stay night", a fictional character. Therefore, in real life, Yao Ming is taller than Saber.

        \\ \hline

    % Exp 
        \multicolumn{4}{|l|}{\textbf{Input:} Which one has a larger registered capital, Dongwu Securities or Xingye Securities?} \\ \hline 
        Model & Search Code & Retrieved Knowledge & Answer  \\ 
    \Xhline{0.8pt}
        % Model
        GPT-4  & 
        % Code
        def search(): \newline  {\color{white}sp} messages = '' \newline  {\color{white}sp} ew\_capital, msg = find\_entity\_or\_value(entity\_aliases = ['Dongwu Securities'], relation\_aliases = ['registered capital', 'capital']) \newline  {\color{white}sp} messages += msg \newline  {\color{white}sp} xy\_capital, msg = find\_entity\_or\_value(entity\_aliases = ['Xingye Securities'], relation\_aliases = ['registered capital', 'capital']) \newline  {\color{white}sp} messages += msg \newline  {\color{white}sp} if ew\_capital and xy\_capital: \newline  {\color{white}sp}  {\color{white}sp} if \wrong{ew\_capital[0] > xy\_capital[0]}: \newline  {\color{white}sp}  {\color{white}sp}  {\color{white}sp} messages += 'Dongwu Securities has a larger registered capital.' \newline  {\color{white}sp}  {\color{white}sp} elif \wrong{ew\_capital[0] < xy\_capital[0]}: \newline  {\color{white}sp}  {\color{white}sp}  {\color{white}sp} messages += 'Xingye Securities has a larger registered capital.' \newline  {\color{white}sp}  {\color{white}sp} else: \newline  {\color{white}sp}  {\color{white}sp}  {\color{white}sp} messages += 'Dongwu Securities and Xingye Securities have the same registered capital.' \newline  {\color{white}sp} return messages \newline  & 
        % Knowledge
        [FROM CNDBPedia][find\_entity\_or\_value(entity\_aliases = ['Dongwu Securities'], relation\_aliases = ['registered capital', 'capital']) -> ] Dongwu Securities, registered capital: 1.5 billion yuan[find\_entity\_or\_value(entity\_aliases = ['Xingye Securities'], relation\_aliases = ['registered capital', 'capital']) -> ] Xingye Securities, registered capital: 6.697 billion yuan (2018)Xingye Securities has a larger registered capital. &
        % Answer
        \textbf{W/ KB} \newline 
        Xingye Securities has a larger registered capital. \newline  \newline 
        \textbf{W/o KB} \newline 
         \avoid{I'm sorry, I can't provide real-time company registered capital information}, as these data may change over time. It is recommended that you directly visit the official websites of these two companies or contact their customer service to get the most accurate information.  \\
        \hline

        % Model
        ChatGPT  & 
        % Code
       def search(): \newline  {\color{white}sp} messages = '' \newline  {\color{white}sp} eastmoney\_capital, msg = find\_entity\_or\_value(entity\_aliases = ['Dongwu Securities'], relation\_aliases = ['registered capital']) \newline  {\color{white}sp} messages += msg \newline  {\color{white}sp} xingye\_capital, msg = find\_entity\_or\_value(entity\_aliases = ['Xingye Securities'], relation\_aliases = ['registered capital']) \newline  {\color{white}sp} messages += msg \newline  {\color{white}sp} return messages \newline 
       & 
        % Knowledge
        [FROM CNDBPedia][find\_entity\_or\_value(entity\_aliases = ['Dongwu Securities'], relation\_aliases = ['registered capital']) -> ] Dongwu Securities, registered capital: 1.5 billion yuan  &
        % Answer
        \textbf{W/ KB} \newline 
        \wrong{(Error when parsing LLM response into json format.)}. \newline  \newline 
        \textbf{W/o KB} \newline 
       \wrong{Dongwu Securities has a larger registered capital}. According to public data, the registered capital of Dongwu Securities is \wrong{RMB 15 billion}, while the registered capital of Xingye Securities is \wrong{RMB 10 billion}.

        \\ \hline
     
    \Xhline{0.8pt}
    \end{tabular}}
    \caption{ The detailed results of comparison queries on CN-DBPedia.  }
    \label{tab:detail_1}
    
\end{table*}

\end{document}